\PassOptionsToPackage{table}{xcolor}
\PassOptionsToPackage{hidelinks}{hyperref}
\documentclass[11pt]{article}

\usepackage[final]{acl}

\usepackage{times}
\usepackage{latexsym}

\usepackage[T1]{fontenc}
\usepackage[utf8]{inputenc}
\usepackage{microtype}
\usepackage{inconsolata}
\usepackage{graphicx}
\usepackage{booktabs}
\usepackage{amsmath}

\usepackage{xurl}
\usepackage[hidelinks]{hyperref}
\usepackage{enumitem}
\usepackage{threeparttable}
\usepackage{pifont}
\usepackage{amssymb}
\usepackage{multirow}
\usepackage{makecell}

\usepackage[table]{xcolor}
\usepackage{tcolorbox}
\tcbset{colback=white, colframe=black, boxrule=0.6pt, arc=1pt, left=6pt, right=6pt, top=6pt, bottom=6pt}
\definecolor{lightpurple}{RGB}{230, 224, 244}
\definecolor{lightorange}{RGB}{255, 236, 214}

\usepackage{pifont}

\usepackage{cuted}
\usepackage{caption}

\usepackage{booktabs}

\newcommand{\BENCH}{\textsc{MiraMind}}
\newcommand{\MODEL}{Mindora}

\usepackage{amssymb}

\title{
\protect\raisebox{\dimexpr\ht\strutbox-1.2em\relax}{%
  \protect\includegraphics[height=2em]{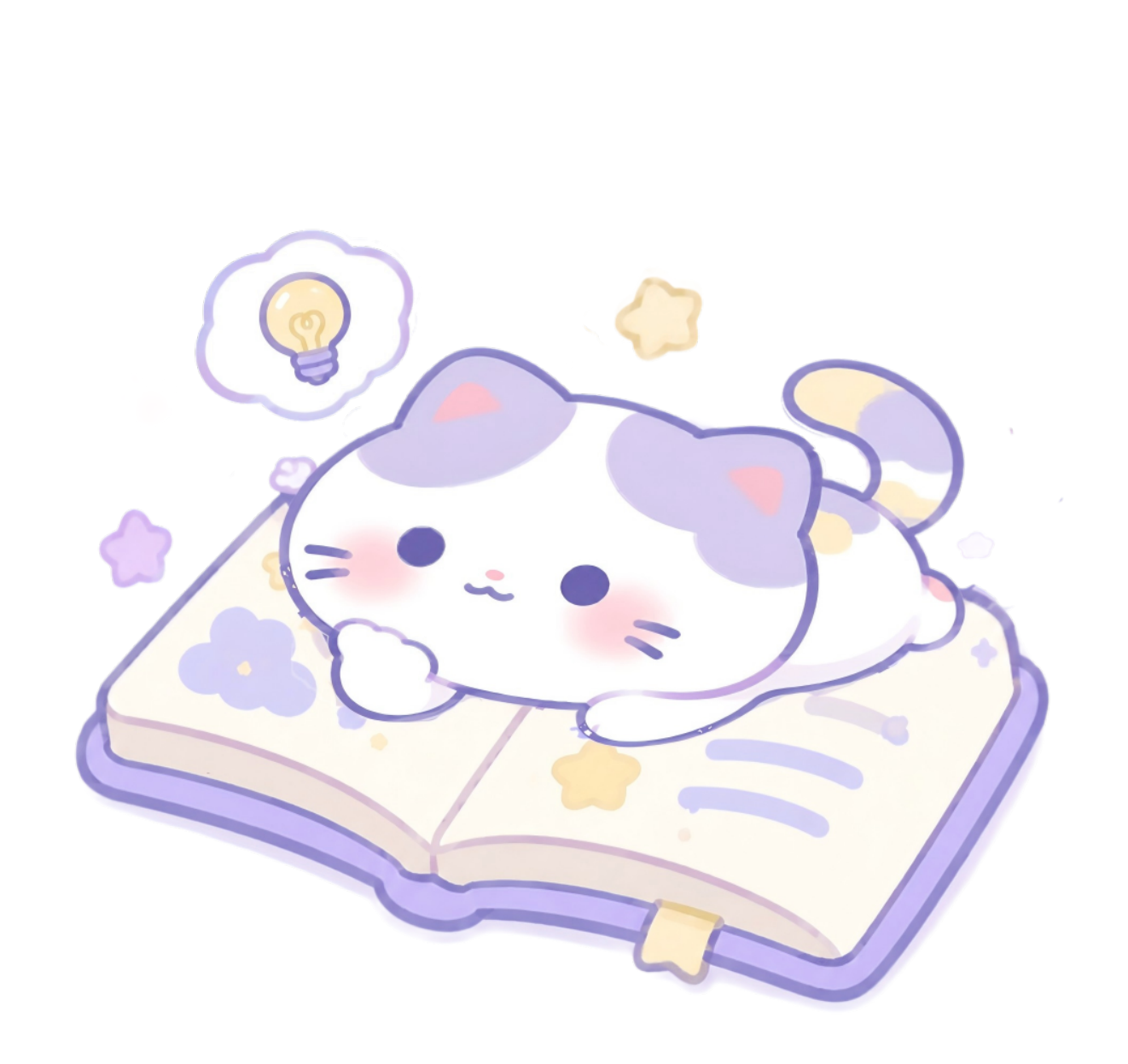}%
}
{\BENCH: Benchmarking Reliable Mental Health Reasoning beyond Answer Accuracy}
}

\author{%
Mengxi Xiao$^{1,2}$, Kailai Yang$^{3}$, Pengde Zhao$^{4}$, Enze zhang$^{1,2}$, Ziyan Kuang$^{1,2}$, Zhiwei Liu$^{3}$,\\ \bf
Weiguang Han$^{4}$, Shu Liao$^{2}$, Lianting Huang$^5$, Guojun Xiong$^6$, Victor Gutierrez Basulto$^7$,\\ \bf
Jinpeng Hu$^8$, Min Peng$^{1,2}$, Qianqian Xie$^{1,2}$\thanks{Corresponding Author. Email: xieq@whu.edu.cn}, Sophia Ananiadou$^{1,3}$
  \\
  $^1$School of Artificial Intelligence, Wuhan University; 
  $^2$ Center for Language and Information \\ Research, Wuhan University; 
  $^3$ University of Manchester;
  $^4$School of Computer Science, 
  \\
  Wuhan University; 
  $^5$Mount Holyoke College;
  $^6$Shanghai Jiao Tong University;
  \\
  $^7$Cardiff University; 
  $^8$Hefei University of Technology
}

\begin{document}
\maketitle
\begin{abstract}
Mental-health reasoning with large language models (LLMs) is an evidence-constrained judgment problem: models must transform limited, subjective, and often ambiguous evidence into interpretations, decisions, or claims whose specificity, certainty, severity, and actionability remain warranted. Existing benchmarks mainly evaluate specific clinical roles or final answers, leaving the reliability of explicit reasoning trajectories under-specified. We introduce \BENCH{}, a unified benchmark spanning six task families and 13 datasets across appraisal, diagnosis, intervention, abstraction, and verification. \BENCH{} evaluates both task outcomes and the trajectories connecting evidence to judgment through usability, logical structure, and informational contribution. Evaluating 20 LLMs reveals a restraint gap, in which the specificity or certainty of model judgments exceeds what limited evidence supports. We further train \MODEL{}, an 8B model that targets evidence-to-judgment transitions through hard-case supervision, structured trajectory rewriting, and consistency-aware optimization. \MODEL{} achieves the best average rank on \BENCH{}, improves over its backbone across all six task families, and produces more balanced reasoning trajectories. These results show that \BENCH{} can expose shared restraint failures and evaluate whether targeted post-training improves evidence-constrained mental-health reasoning.
\end{abstract}

\section{Introduction}
\begin{figure}[ht]
    \centering
    \includegraphics[width=\linewidth]{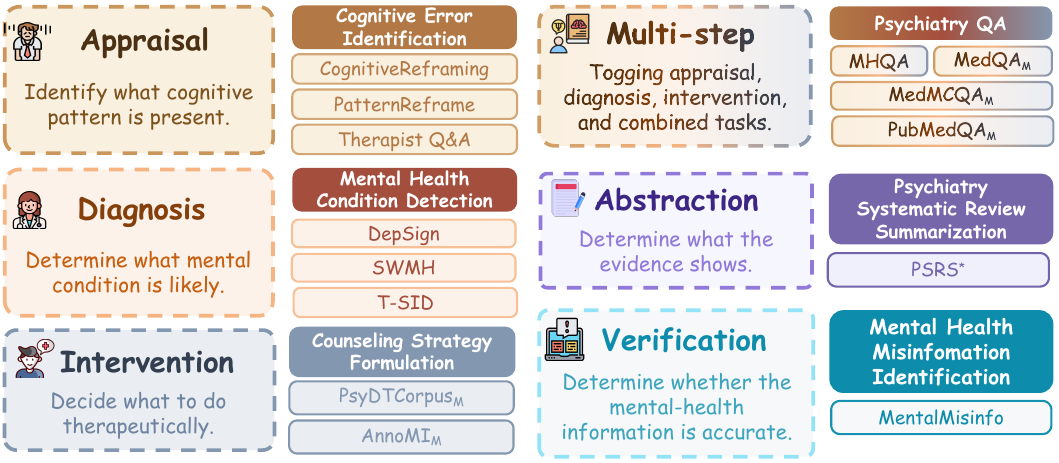}
    \caption{Tasks and datasets included in \BENCH.}
    \label{fig:main}
\end{figure}
% Mental health disorders affect hundreds of millions of people worldwide and remain a leading contributor to disability, social inequality, and unmet clinical needs~~\cite{WHO2025,Patel2018LancetCommission,GBD2019MentalDisorders2022}. At the same time, large language models (LLMs) have shown strong potential for supporting mental-health assessment and reasoning, ranging from interpreting subjective self-reports to assisting clinicians and counselors in decision-making and treatment planning~\citep{healme,singh2024deciphering,yang2024mentallama,annomi2022,qiu2024interactive,moodangels}. However, deploying LLMs in such high-stakes scenarios requires more than producing fluent outputs or superficially correct answers: effective mental-health support depends on reasoning that is transparent, coherent, and context-grounded, reflecting how humans integrate abstract psychological constructs with case-specific evidence. When models misinterpret narratives, rely on incomplete reasoning, or introduce ungrounded assumptions, they may amplify anxiety, provide misleading feedback, and ultimately undermine trust. These risks highlight an urgent need for evaluation frameworks that quantify not only task correctness but also the reliability of reasoning trajectories.

\begin{table*}[t]
\centering
\renewcommand{\arraystretch}{1.25}
\caption{Comparison between \BENCH{} and existing mental-health reasoning benchmarks. 
\BENCH{} differs from prior work by jointly covering heterogeneous mental-health reasoning functions and evaluating reasoning trajectories beyond final-answer accuracy.}
\label{tab:benchmark_comparison}
\resizebox{\linewidth}{!}{
\begin{tabular}{llccccccc}
\hline
\textbf{Benchmark} 
& \textbf{Main Focus}
& \makecell[c]{\textbf{Appraisal}}
& \makecell[c]{\textbf{Diagnosis}}
& \makecell[c]{\textbf{Intervention}}
& \makecell[c]{\textbf{Abstraction}}
& \makecell[c]{\textbf{Verification}\\\textbf{/ Safety}}
& \makecell[c]{\textbf{Multi-task}\\\textbf{Coverage}}
& \makecell[c]{\textbf{Trajectory}\\\textbf{ Eval.}} \\
\hline

\multicolumn{9}{c}{\textit{Psychiatric diagnosis and clinical decision-making benchmarks}} \\ \hline

MentalBench~\citep{song2026mentalbench}
& DSM-5 diagnostic decision-making
& \ding{55} & \ding{51} & \ding{55} & \ding{55} & \ding{55}
& \ding{55} & \ding{55} \\

MentalDx Bench~\citep{sun2026mentalseekdx}
& Real-world psychiatric diagnosis from EHRs
& \ding{55} & \ding{51} & \ding{55} & \ding{55} & \ding{55}
& \ding{55} & \ding{55} \\

LingxiDiagBench~\citep{xu2026lingxidiagbench}
& Multi-turn Chinese psychiatric consultation and diagnosis
& \ding{55} & \ding{51} & \ding{51} & \ding{55} & \ding{55}
& \ding{51} & \ding{55} \\

PsychiatryBench~\citep{fouda2026psychiatrybench}
& Psychiatry textbook/casebook QA
& \ding{55} & \ding{51} & \ding{51} & \ding{55} & \ding{55}
& \ding{51} & \ding{55} \\

MENTAT~\citep{lamparth2025moving}
& Real-world psychiatric decision-making
& \ding{55} & \ding{51} & \ding{51} & \ding{55} & \ding{51}
& \ding{51} & \ding{55} \\

\hline
\multicolumn{9}{c}{\textit{Counseling and psychotherapy competence benchmarks}} \\ \hline

CBT-Bench~\citep{zhang2025cbtbench}
& CBT knowledge, cognitive model understanding, and response generation
& \ding{51} & \ding{55} & \ding{51} & \ding{55} & \ding{55}
& \ding{51} & \ding{55} \\

CounselingBench~\citep{nguyen2025counselingbench}
& Counseling competency examination
& \ding{51} & \ding{51} & \ding{51} & \ding{55} & \ding{51}
& \ding{51} & \ding{55} \\

CounselBench~\citep{li2025counselbench}
& Counseling response quality and adversarial safety
& \ding{55} & \ding{55} & \ding{51} & \ding{55} & \ding{51}
& \ding{55} & \ding{55} \\

MindEval~\citep{pombal2025mindeval}
& Multi-turn mental-health support interaction
& \ding{55} & \ding{55} & \ding{51} & \ding{55} & \ding{51}
& \ding{55} & \ding{55} \\

\hline
\multicolumn{9}{c}{\textit{Safety, crisis, and ethical reasoning benchmarks}} \\ \hline

EthicsMH~\citep{kasu2025ethicsmh}
& Ethical reasoning in mental-health AI
& \ding{55} & \ding{55} & \ding{55} & \ding{55} & \ding{51}
& \ding{55} & \ding{55} \\

\makecell[l]{Mental Health Crisis Benchmark\\ \citep{arnaiz2026between}}
& Crisis detection and crisis response
& \ding{55} & \ding{51} & \ding{51} & \ding{55} & \ding{51}
& \ding{55} & \ding{55} \\

\hline
\BENCH{} (ours)
& Reliable mental-health reasoning beyond answer accuracy
& \ding{51} & \ding{51} & \ding{51} & \ding{51} & \ding{51}
& \ding{51} & \ding{51} \\
\hline
\end{tabular}
}
\end{table*}

When people ask LLMs about mental health, they rarely ask a clean question with a checkable answer. They bring a story and ask the model to help decide what it means, whether a thought should worry them, whether a feeling suggests risk, what response would help, or whether a piece of advice should be trusted~\citep{healme,singh2024deciphering,yang2024mentallama,qiu2024interactive,moodangels}. This is where today's reasoning LLMs face a problem that their gains on mathematics, coding, and scientific problem solving do not resolve. In those domains, post-training can often reward reasoning because the final answer is externally verifiable~\citep{zelikman2022star,lightman2023let,deepseek2025r1}.

We characterize this challenge as \emph{evidence-constrained judgment}: a model must transform limited, subjective, and often ambiguous evidence into an interpretation, decision, or claim whose specificity, certainty, severity, and actionability remain warranted by that evidence. A \emph{restraint failure} occurs when the strength of the model's judgment exceeds what the available evidence supports, such as when distress is converted into a specific diagnosis or an ambiguous situation into a confident intervention recommendation.

In mental health, intermediate interpretations, including what the model treats as a symptom, risk signal, appropriate response, or credible evidence, may shape the user's self-understanding and subsequent action before external verification is possible. Recent lawsuits and investigations have made these stakes visible\footnote{We refer to these cases as public allegations and investigations, not as established causal findings. Examples include public reports on lawsuits or investigations concerning chatbot interactions in adolescent suicide (\href{https://time.com/7312484/chatgpt-openai-suicide-lawsuit/}{link}), delusional escalation before murder-suicide (\href{https://apnews.com/article/97fd7da31c0fa08f3d3ea9efd6713151}{link}), and school or campus shooting planning (\href{https://www.pbs.org/newshour/nation/lawsuit-accuses-chatgpt-of-helping-gunman-plan-fsu-mass-shooting}{link}).}.

Recent mental-health benchmarks have moved beyond early social-media classification. Psychiatric benchmarks focus on diagnosis and clinical decision-making~\citep{song2026mentalbench,sun2026mentalseekdx,xu2026lingxidiagbench,fouda2026psychiatrybench,lamparth2025moving}; counseling-oriented benchmarks assess CBT assistance, counseling competence, and response quality~\citep{zhang2025cbtbench,nguyen2025counselingbench,li2025counselbench,pombal2025mindeval}; and safety-focused resources examine crisis handling and ethical reasoning~\citep{arnaiz2026between,kasu2025ethicsmh}. While clinically valuable, these benchmarks usually test a specific role or endpoint, as summarized in Table~\ref{tab:benchmark_comparison}. Final-answer metrics reveal whether a model reaches the expected endpoint, but not whether its judgment is appropriately supported. A correct diagnosis, strategy, or risk label may still rely on unsupported assumptions, excessive certainty, or unnecessary pathologization. Reliable evaluation must therefore inspect not only the endpoint, but also how evidence is transformed into judgment.

% To address this gap, we introduce \BENCH, a unified benchmark for evaluating whether LLMs can reason reliably in mental-health settings. Unlike prior evaluations that mainly test isolated knowledge, prediction, or final-answer selection, \BENCH{} is built to expose how models move from limited mental-health evidence to cautious judgments. We first curate 13 datasets across the main forms of evidence that appear in mental-health use, including personal posts, counseling dialogues, psychiatry questions, systematic-review abstracts, and online mental-health claims. This requires more than aggregation, as we refine existing resources, construct task-specific versions of medical and counseling datasets, and introduce new expert-annotated evidence summarization data. We then organize these datasets into six task families that make different parts of the judgment process testable, including interpreting problematic thought patterns, identifying possible mental-health conditions, selecting counseling responses, answering psychiatry questions, synthesizing clinical evidence, and verifying misleading claims. Finally, we evaluate not only whether the final answer matches a reference, but also whether the reasoning trajectory is usable, logically coherent, and informative, allowing \textsc{MiraMind} to distinguish models that reach the expected endpoint\nb{V: This sentence is incomplete. They reached it ``by chance''?} from those that reach it through grounded and reliable reasoning.

To address this gap, we introduce \BENCH{}, a unified benchmark that operationalizes evidence-constrained mental-health judgment across six task families and evaluates both final decisions and the explicit trajectories connecting evidence to those decisions. These task families instantiate the same underlying problem under different forms of evidence and different consequences of judgment. Appraisal and condition detection test how models interpret sparse personal evidence; intervention and psychiatry QA test how interpretations are converted into decisions; abstraction and verification test how models integrate external evidence before accepting or communicating a claim. The contribution is therefore not the mere aggregation of 13 datasets, but a common evaluation structure that makes evidence-to-judgment failures comparable across otherwise isolated mental-health tasks. This structure combines refined existing resources, task-specific versions of medical and counseling datasets, and a new expert-annotated evidence-summarization dataset (Table~\ref{tab:dataset_statistics}). \BENCH{} evaluates final-answer performance together with trajectory usability, logical structure, and informational contribution.

We evaluate 20 LLMs on \BENCH{} to examine whether restraint failures recur across models and task families. When such failures appear systematically across different evidence-to-judgment settings, they form what we call a \emph{restraint gap}. Our results reveal this shared pattern: strong models often recognize familiar labels or produce plausible explanations, yet still overread weak cues, introduce generic clinical assumptions, or provide rationales that do not support the strength of their conclusions. These failures become visible because \BENCH{} varies both the evidence source and the judgment required while evaluating trajectories in addition to endpoints.

We then test whether \BENCH{} captures a shared failure in evidence control. If this failure is transferable across task families, post-training that explicitly targets evidence-to-judgment transitions should produce consistent cross-task improvements. To examine this hypothesis, we train \MODEL{}, an 8B model using hard-case supervision, structured trajectory rewriting, and consistency-aware optimization. \MODEL{} achieves the highest overall rank among the evaluated systems and produces more balanced reasoning trajectories. Its gains across multiple task families support the interpretation that targeted post-training improves a transferable control capability rather than fitting a single dataset.

Our contributions are summarized as follows: \textbf{(i)} We introduce \BENCH{}, which formulates reliable mental-health reasoning as evidence-constrained judgment across six task families and 13 datasets. \textbf{(ii)} We develop a trajectory-level evaluation protocol that audits how evidence is transformed into judgment through usability, logical structure, and informational contribution. \textbf{(iii)} We evaluate 20 LLMs and identify recurring restraint failures that form a shared cross-task restraint gap; through \MODEL{}, we further provide evidence that targeted post-training can improve this control across heterogeneous task families.

\section{\BENCH{} Benchmark}
\label{sec:benchmark}

\begin{table*}[t]
\centering
\renewcommand{\arraystretch}{1.25}
\caption{Task and dataset summary of \BENCH{}. Datasets with $_M$ are processed in this work. *PSRS is newly curated and expert-annotated in this work. The \BENCH{} Adaptation column distinguishes prompt and format standardization, task reformulation, subset curation, and new expert annotation. The \textit{recall} metric for PSRS measures the coverage of expert-annotated scoring points.}
\label{tab:dataset_statistics}
\begin{threeparttable}
\resizebox{\linewidth}{!}{
\begin{tabular}{lccclc}
\hline
\textbf{Dataset}
& \textbf{Train/Valid/Test}
& \textbf{Source}
& \textbf{Annotation}
& \textbf{\BENCH{} Adaptation}
& \textbf{Metric}
\\
\hline

\multicolumn{6}{c}{\textit{Cognitive Error Identification}} \\ 
\hline
\makecell[l]{CognitiveReframing\\ \citep{sharma2023cognitive}}
& 694/173/284
& synthetic, MHA
& Human
& \makecell[l]{Standardized prompts, answer formats, and evaluation;\\ original labels preserved}
& Micro-F1 \\

\makecell[l]{PatternReframe\\ \citep{Mounica2023Training}}
& 1214/661/300
& synthetic
& Human
& \makecell[l]{Standardized prompts, answer formats, and evaluation;\\ original labels preserved}
& Micro-F1 \\

\makecell[l]{Therapist Q\&A\\ \citep{shreevastava2021detecting}}
& 1508/409/507
& \makecell[c]{counseling dialogues (real)}
& Human
& \makecell[l]{Reformulated as utterance-level cognitive-error\\ identification; original labels preserved}
& Micro-F1 \\

\hline
\multicolumn{6}{c}{\textit{Mental Health Condition Detection}} \\ 
\hline
\makecell[l]{DepSign\\ \citep{depsign}}
& 1113/300/600
& Reddit
& weak supervision
& \makecell[l]{Standardized prompts, answer formats, and evaluation;\\ original labels preserved}
& Micro-F1 \\

\makecell[l]{SWMH\\ \citep{ji2022suicidal}}
& 1097/300/500
& Reddit
& weak supervision
& \makecell[l]{Standardized prompts, answer formats, and evaluation;\\ original labels preserved}
& Micro-F1 \\

\makecell[l]{T-SID\\ \citep{ji2022suicidal}}
& 1080/300/942
& Twitter
& weak supervision
& \makecell[l]{Standardized prompts, answer formats, and evaluation;\\ original labels preserved}
& Micro-F1 \\

\hline
\multicolumn{6}{c}{\textit{Counseling Strategy Formulation}} \\ 
\hline
PsyDTCorpus$_M$
& 1105/300/-
& \makecell[c]{counseling dialogues (syn)}
& Human
& \makecell[l]{Compressed client utterances into case summaries;\\ extracted 13 strategy labels; expert reviewed}
& Jaccard \\

AnnoMI$_M$
& -/-/133
& \makecell[c]{counseling dialogues (real)}
& Human
& \makecell[l]{Compressed client utterances into case summaries;\\ extracted 13 strategy labels; expert reviewed}
& Jaccard \\

\hline
\multicolumn{6}{c}{\textit{Psychiatry QA}} \\ 
\hline
\makecell[l]{MHQA\\ \citep{racha2025mhqa}}
& 1123/300/717
& PubMed
& Human/LLM
& \makecell[l]{Standardized prompts, answer formats, and evaluation;\\ original labels preserved}
& Micro-F1 \\

MedQA$_M$
& 415/58/121
& Medical exams
& Human
& \makecell[l]{Manually selected psychiatry-related items;\\ original answer labels preserved}
& Micro-F1 \\

MedMCQA$_M$
& 287/120/446
& Medical exams
& Human
& \makecell[l]{Manually selected psychiatry-related items;\\ original answer labels preserved}
& Micro-F1 \\

PubMedQA$_M$
& -/-/89
& PubMed
& Human
& \makecell[l]{Manually selected psychiatry-related items;\\ original answer labels preserved}
& Micro-F1 \\

\hline
\multicolumn{6}{c}{\textit{Psychiatry Systematic Review Summarization}} \\ 
\hline
*PSRS
& -/-/108
& Cochrane Library
& Human
& \makecell[l]{Newly curated summarization task; expert scoring\\ points annotated, calibrated, and cross-checked}
& Recall \\

\hline
\multicolumn{6}{c}{\textit{Mental Health Misinformation Identification}} \\ 
\hline
\makecell[l]{MentalMisinfo\\ \citep{nguyen2025supporters}}
& 123/31/130
& YouTube, BitChute
& Human
& \makecell[l]{Standardized prompts, answer formats, and evaluation;\\ original labels preserved}
& Macro-F1 \\

\hline
\end{tabular}
}
\end{threeparttable}
\end{table*}

\BENCH{} is designed to evaluate reliable mental-health reasoning across the types of evidence, judgments, and explanations that arise in user-facing mental-health settings. It organizes six task families under five core aspects: appraisal, diagnosis, intervention, abstraction, and verification, with psychiatry question answering integrating multiple aspects within a single case. Although these tasks differ in their inputs and outputs, they require the same underlying control: the strength of an interpretation, decision, or claim must remain proportional to the evidence available.

Our benchmark construction follows three principles. First, each task must involve a consequential mapping from mental-health evidence to an interpretation, decision, or claim. Second, each dataset should preserve the incompleteness, ambiguity, or uncertainty of the original evidence whenever possible. Third, evaluation should examine both the task endpoint and the explicit trajectory through which evidence is transformed into judgment. Table~\ref{tab:dataset_statistics} distinguishes resources that are standardized, task-adapted, manually subsetted, or newly curated, and summarizes the corresponding annotation effort. Exact licenses, access conditions, and redistribution constraints are reported in Appendix~\ref{append:benchmark_details}.

\subsection{Task Definition and Dataset Curation}
\label{sec:task_dataset}

\paragraph{Appraisal: Cognitive-Pattern Reasoning.}
% Appraisal tests whether a model can identify the maladaptive cognitive pattern expressed in a client statement. This ability is necessary because user-facing mental-health support often begins before formal diagnosis: the model must interpret how a person frames an event, whether the thought reflects overgeneralization, catastrophizing, personalization, mind-reading, or another cognitive distortion.
Appraisal examines how models transform short personal statements into cognitive-pattern interpretations. Because these statements provide limited context, reliable judgment requires identifying the best-supported pattern without treating a tentative cue as a definitive psychological characterization.

\noindent\textit{Task definition:}
Given a short situation, such as a thought pair or a client utterance, the model selects the most appropriate cognitive-error label from the taxonomy defined by the source dataset. The task is formulated as multi-class classification and evaluated with Micro-F1.

\noindent\textit{Dataset curation:}
We instantiate this task using CognitiveReframing, PatternReframe, and Therapist Q\&A. These datasets differ in source and taxonomy: CognitiveReframing combines simulated negative thoughts with Mental Health America self-reports annotated by trained mental-health professionals; PatternReframe contains crafted statements labeled by independent raters; Therapist Q\&A provides naturally occurring distorted statements from therapist-client interactions. Because the taxonomies are not identical, we preserve each dataset's original label set and prompt definitions rather than forcing a lossy unified taxonomy. This design evaluates whether models can adapt to related but non-identical appraisal schemas. Further details are in Appendix~\ref{append:appraisal_dataset_details}.

\paragraph{Diagnosis: Mental-Condition Reasoning.}
% Diagnosis evaluates whether a model can infer possible mental-health conditions from short, subjective user narratives while avoiding over-pathologizing. This differs from psychiatric EHR diagnosis: inputs are informal posts or tweets, evidence is incomplete, and the correct behavior is calibrated condition detection rather than exhaustive clinical labeling.
Diagnosis examines how models transform short, subjective user narratives into condition or severity judgments. Unlike psychiatric diagnosis based on comprehensive clinical records, these inputs are informal and incomplete. Models must therefore identify the most supported label without converting weak evidence into a definitive disorder or severe-risk judgment.

\noindent\textit{Task definition:}
Given a user post, the model predicts the most appropriate condition or severity label according to the dataset-specific label space. DepSign evaluates depression severity, while SWMH and T-SID evaluate multiple mental-health condition categories. We evaluate classification performance with Micro-F1.

\noindent\textit{Dataset curation:}
We use DepSign, SWMH, and T-SID, covering Reddit and Twitter-style language. These datasets provide naturalistic mental-health expressions, including noisy, incomplete, and emotionally loaded narratives. We retain the original task labels to preserve dataset-specific clinical assumptions, while standardizing prompts and answer formats across datasets.

\paragraph{Intervention: Therapeutic-Action Reasoning.}
% Intervention tests whether a model can decide what therapeutic action is appropriate in a counseling context. This is distinct from generating an empathetic reply: the model must identify the counseling function needed by the client situation, such as clarification, reflection of feeling, summarizing, confrontation, encouragement, interpretation, or guidance.
Intervention examines how models transform counseling evidence into therapeutic-action decisions. The task requires identifying the counseling function supported by the client situation, rather than generating a generally empathetic response. Because several strategies may be plausible, models must avoid inferring unsupported needs or selecting an intervention stronger than the available case evidence warrants.

\noindent\textit{Task definition:}
Given a concise client case derived from a counseling dialogue, the model selects one or more appropriate counseling strategies from 13 commonly taught intervention types. Because multiple strategies may be acceptable for a single counseling turn, we evaluate predictions with Jaccard similarity against the reference strategy set.

\noindent\textit{Dataset construction:}
We construct two processed datasets, PsyDTCorpus$_M$ and AnnoMI$_M$. Client utterances are compressed into concise case summaries, and counselor responses are annotated to extract strategy labels as reference answers. All constructed items are manually reviewed by an expert counselor with over ten years of clinical experience. To avoid evaluating on the same distribution used for training, PsyDTCorpus$_M$ is used only for training because of its partially synthetic nature, whereas AnnoMI$_M$ is reserved for testing and is derived from real motivational-interviewing demonstrations. This source separation enables a cross-source test of whether intervention reasoning transfers beyond the training distribution. Construction details are in Appendix~\ref{append:intervention_dataset_detail}.

\paragraph{Multi-step: Psychiatry Question Answering.}
% Multi-step reasoning evaluates whether models can integrate symptom interpretation, diagnostic inference, intervention selection, and evidence-based medical knowledge within a single case. This setting approximates clinician-side reasoning more closely than single-label appraisal or condition detection.
Psychiatry question answering examines how models integrate symptoms, case details, medical knowledge, and possible interventions into a final decision. Although these questions often have a verifiable answer, reliable reasoning still requires that the selected option follow from the presented evidence and relevant knowledge rather than unsupported clinical assumptions.

\noindent\textit{Task definition:}
Given a psychiatry-related question or clinical vignette with candidate answers, the model selects the correct answer and provides a reasoning trajectory. We evaluate final-answer correctness with Micro-F1.

\noindent\textit{Dataset curation:}
We use MHQA and construct psychiatry-focused subsets from MedQA, MedMCQA, and PubMedQA, denoted with the $_M$ suffix\footnote{Brief introductions to the original datasets are provided in Appendix~\ref{append:qa_raw_data_info}.}. For the broad medical QA datasets, we manually extract items involving psychiatric disorders, psychopharmacology, therapeutic management, mental-health symptoms, or psychiatric research evidence. This processing isolates mental-health reasoning from general medical QA and ensures that the task evaluates psychiatry-relevant multi-step judgment rather than broad biomedical knowledge alone.

\paragraph{Abstraction: Evidence-Based Psychiatry Summarization.}
% Abstraction evaluates whether a model can synthesize psychiatric research evidence into clinically meaningful conclusions. This ability is important because mental-health users and practitioners often encounter claims grounded in studies, reviews, or guidelines, where the core challenge is not choosing a label but determining what the evidence actually supports.
Abstraction examines how models transform psychiatric research evidence into clinically meaningful claims. Models must preserve the direction, magnitude, and certainty of the reported findings, ensuring that the resulting summary does not state a stronger or more definite conclusion than the underlying evidence supports.

\noindent\textit{Task definition:}
Given a structured abstract from a psychiatry-related Cochrane systematic review, the model generates the Main Results summary, focusing on the direction, magnitude, and certainty of the reported findings. Unlike extractive summarization, the task requires evidence abstraction: models must convert statistical and methodological details into a concise clinical interpretation.

\noindent\textit{Dataset construction:}
We introduce Psychiatry Systematic Review Summarization (PSRS), a newly curated and expert-annotated generation task. We collect 115 structured abstracts from the Cochrane Library in the mental-health domain, preserving review identifiers for traceability and using only publicly available abstract text. For annotation, LLM-generated Main Results summaries are evaluated by two psychiatrists using a 0--2 rubric: accurate and complete, partially correct or incomplete, and unsupported or substantially incorrect. A pilot calibration phase is used to align judgment thresholds, and a cross-check subset is used for quality assurance. The benchmark reports recall over expert-annotated scoring points, measuring whether a model covers the clinically relevant evidence conclusions. Construction and annotation details are in Appendix~\ref{append:abstraction_dataset_construction}.

\paragraph{Verification: Mental-Health Misinformation Detection.}
% Verification evaluates whether a model can judge the accuracy of mental-health information in noisy online environments. This aspect is necessary because many real user interactions involve asking whether advice, claims, or social-media content should be trusted.
Verification examines how models use noisy online evidence to decide whether a mental-health claim should be accepted or rejected. Reliable judgment requires distinguishing persuasive presentation from evidential support and avoiding conclusions that exceed the information available in the content and its context.

\noindent\textit{Task definition:}
Given video-derived textual content, including titles, hashtags, transcripts, and contextual metadata when available, the model predicts whether the mental-health claim is accurate or misleading. We evaluate this binary classification task with Macro-F1 to account for class imbalance.

\noindent\textit{Dataset curation:}
We use MentalMisinfo, which contains mental-health-related video content from platforms such as YouTube Shorts and BitChute. The materials are filtered, transcribed, and manually annotated as accurate or misleading. This task extends \BENCH{} beyond clinical decision-making by testing evidence-based verification under informal and persuasive online discourse.

\subsection{Reasoning Trajectory Evaluation}
\label{sec:trajectory_eval}

\BENCH{} evaluates explicit reasoning trajectories in addition to final answers. We score final answers with task-specific metrics and assess trajectories along three dimensions: usability, logical structure, and informational contribution. These dimensions capture observable forms of trajectory-level restraint failure: excessive terminology and verbosity may obscure weak evidential support; unsupported links, irrelevant clinical templates, and contradictions indicate unstable evidence-to-judgment transitions; and redundant steps or missing qualifications leave judgments insufficiently constrained.

Operationally, usability is measured by Terminology Parsimony and Step Adequacy; logical structure by Logical Coherence, Irrelevance Rate, and Contradiction Rate; and informational contribution by Informativeness. An LLM-as-judge segments each trajectory into reasoning content units and labels their relations to the input and preceding units. We validate the protocol on 100 trajectories independently annotated by two trained evaluators. Appendix~\ref{append:chain_eval_detail} reports evaluator qualifications, calibration procedures, inter-annotator agreement, LLM-human agreement, and a representative disagreement case.

These metrics audit the structure and grounding of explicit rationales, but do
not by themselves establish clinical correctness, safety, or faithfulness to
the model's latent computation.

\section{\MODEL{}: Targeted Post-Training for Evidence-Constrained Judgment}
\label{sec:Mindora}

If \BENCH{} captures a shared evidence-control failure rather than a collection of unrelated dataset-specific weaknesses, then post-training explicitly targeted at this failure should improve performance across task families with different evidence forms and judgment outputs. We therefore train \MODEL{}, based on Qwen3-8B, as an intervention test of this benchmark hypothesis. The goal is not to introduce a standalone clinical system, but to examine whether a compact model can learn to better control the transformation from limited evidence to proportionate and internally consistent judgments.

\MODEL{} follows a trajectory-oriented post-training pipeline with three main components. First, we construct supervision from hard cases where a base model fails under zero-shot prompting. These instances are less likely to be solved through surface cues and therefore provide higher-signal examples for learning evidence-sensitive distinctions. Second, we generate and rewrite reasoning trajectories into concise, non-backtracking explanations with explicit evidence, intermediate inferences, and final judgments. This structured rewriting makes the evidence--inference--judgment relation more explicit and reduces unsupported transitions, omitted qualifiers, and generic reasoning templates. Third, we train the model with a hybrid SFT--RL objective using a consistency-aware reward that combines format validity, length validity, internal consistency, and task-specific correctness. This component discourages trajectories whose intermediate reasoning is disconnected from, or inconsistent with, the final judgment.

The full training pipeline, including hard-case filtering, verifier-guided trajectory search, structured rewriting, and reward construction, is provided in Appendix~\ref{append:mindora_training}. Section~\ref{sec:experiments} evaluates whether this intervention produces the cross-task improvements predicted by the benchmark hypothesis.

\section{Experiments}
\label{sec:experiments}

%主实验结果
\begin{table*}[htbp]
\centering
\renewcommand{\arraystretch}{1.3}
% \caption{Experimental results on \BENCH, where Setting 1 (S1) stands for hard-sample filtering; Setting 2 (S2) stands for structured rewriting; Setting 3 (S3) stands for Consistency Reward.}
\caption{Main benchmark results on \BENCH{}. Results are grouped by task family. Avg$_1$, Avg$_2$, and Avg$_4$ denote average scores for appraisal, diagnosis, and multi-step psychiatry QA, respectively; Avg$_{all}$ is the average across all evaluated datasets. Higher scores indicate better performance. Mindora w/o S1/S2/S3 are ablation variants, where S1 denotes hard-case filtering, S2 denotes structured rewriting, and S3 denotes the consistency-aware reward.}
\label{tab:main_result}
\resizebox{\linewidth}{!}{% 自动适配表格宽度
\begin{tabular}{l|cccc|cccc|c|ccccc|c|c|cc}
\hline
\textbf{Model} 
& \makecell[c]{\textbf{Cognitive}\\\textbf{Reframing}}
& \makecell[c]{\textbf{Pattern}\\\textbf{Reframe}}
& \makecell[c]{\textbf{Therapist}\\\textbf{Q\&A}}
& \textcolor{brown}{$Avg_{1}$}
& \textbf{DepSign} 
& \textbf{SWMH} 
& \textbf{T-SID} 
& \textcolor{brown}{$Avg_{2}$}
& \textbf{AnnoMI$_{M}$} 
& \textbf{MHQA} 
& \textbf{MedQA$_{M}$} 
& \textbf{MedMCQA$_{M}$} 
& \textbf{PubMedQA$_{M}$} 
& \textcolor{brown}{$Avg_{4}$}
& \textbf{PSRS}
& \makecell[c]{\textbf{Mental}\\\textbf{Misinfo}}
& \textcolor{brown}{$Avg_{all}$} 
& \textcolor{brown}{$Avg_{rank}$} \\
\hline

\multicolumn{19}{c}{\textit{close-source models}} \\ \hline
GPT-o4-mini~\citep{o4_mini} 
& 0.7002 & 0.6607 & 0.4468 & \textcolor{brown}{0.6026}
& 0.3607 & 0.7374 & 0.7515 & \textcolor{brown}{0.6165}
& 0.2086 
& 0.5462 & \cellcolor{lightorange}\textbf{0.8925} & 0.8102 & 0.7118 & \textcolor{brown}{0.7402}
& 0.8650 
& \cellcolor{lightorange}\textbf{0.7781}
& \textcolor{brown}{0.6515} 
& \textcolor{brown}{12}
\\

GPT-4o~\citep{gpt-4o} 
& 0.6791 & 0.6667 & 0.4543 & \textcolor{brown}{0.6000}
& 0.4064 & 0.7500 & 0.7114 & \textcolor{brown}{0.6226}
& 0.3358 
& 0.4531 & 0.6138 & 0.7936 & 0.6949 & \textcolor{brown}{0.6389}
& 0.9065 
& 0.6894 
& \textcolor{brown}{0.6273} 
& \textcolor{brown}{14}
\\

Deepseek-R1~\citep{ds-R1} 
& \cellcolor{lightorange}\textbf{0.7516} & \cellcolor{lightorange}\textbf{0.7069} & 0.4472 & \textcolor{brown}{0.6352}
& 0.4085 & 0.7624 & 0.7218 & \textcolor{brown}{0.6309}
& 0.2020 
& 0.4984 & 0.8608 & {0.8988} & 0.7256 & \textcolor{brown}{0.7459}
& 0.9037 
& 0.5689
& \textcolor{brown}{0.6505} 
& \textcolor{brown}{10}
\\

Deepseek-V3~\citep{ds-v3} 
& 0.6540 & 0.6755 & 0.4131 & \textcolor{brown}{0.5809}
& 0.4211 & 0.7715 & 0.7581 & \textcolor{brown}{0.6502}
& 0.2140 
& 0.4768 & 0.8184 & 0.8696 & 0.6848 & \textcolor{brown}{0.7124}
& 0.9296 
& 0.6156 
& \textcolor{brown}{0.6386} 
& \textcolor{brown}{11}
\\

Qwen-plus~\citep{qwen3} 
& 0.6821 & 0.6395 & 0.4057 & \textcolor{brown}{0.5758}
& 0.4043 & 0.7624 & 0.7662 & \textcolor{brown}{0.6443}
& 0.1866 
& 0.4795 & 0.8242 & 0.8895 & 0.7199 & \textcolor{brown}{0.7283}
& 0.9048
& 0.6018 
& \textcolor{brown}{0.6387} 
& \textcolor{brown}{12}
\\

QwQ-plus~\citep{qwen2.5} 
& 0.6698 & 0.6871 & 0.3698 & \textcolor{brown}{0.5756}
& 0.4085 & 0.7547 & 0.7330 & \textcolor{brown}{0.6321}
& 0.1848 
& 0.4201 & 0.6405 & 0.8354 & 0.7226 & \textcolor{brown}{0.6546}
& 0.9335 
& 0.4838 
& \textcolor{brown}{0.6034} 
& \textcolor{brown}{15}
\\
\hline

\multicolumn{19}{c}{\textit{70B+ open-source models}} \\ \hline
LLaMA-4~\citep{llama4} 
& 0.6588 & 0.6486 & 0.4031 & \textcolor{brown}{0.5702}
& 0.3914 & 0.7310 & 0.7884 & \textcolor{brown}{0.6369}
& 0.2066 
& 0.4684 & 0.7414 & 0.8900 & 0.7239 & \textcolor{brown}{0.7059}
& 0.7634 
& 0.6158 
& \textcolor{brown}{0.6178} 
& \textcolor{brown}{15}
\\

LLaMA-3.3-70B~\citep{llama3} 
& 0.6444 & 0.6301 & 0.3566 & \textcolor{brown}{0.5437}
& 0.4148 & 0.7163 & 0.7157 & \textcolor{brown}{0.6156}
& 0.2023 
& 0.4148 & 0.7712 & 0.8904 & 0.7299 & \textcolor{brown}{0.7016}
& 0.6318
& 0.6894
& \textcolor{brown}{0.6006} 
& \textcolor{brown}{16}
\\

dsdistill-LLaMA-70B~\citep{ds-R1} 
& 0.6667 & 0.6667 & 0.3905 & \textcolor{brown}{0.5746}
& 0.4000 & 0.7516 & 0.6953 & \textcolor{brown}{0.6156}
& 0.2367 
& 0.4404 & 0.7097 & 0.7957 & 0.7452 & \textcolor{brown}{0.6727}
& 0.8914 
& 0.6814 
& \textcolor{brown}{0.6209}
& \textcolor{brown}{14}
\\

Qwen2.5-72B~\citep{qwen2.5} 
& 0.6852 & 0.6207 & 0.3930 & \textcolor{brown}{0.5663}
& 0.3806 & 0.7639 & 0.8136 & \textcolor{brown}{0.6527}
& 0.2796 
& 0.4022 & 0.7449 & 0.8402 & 0.7229 & \textcolor{brown}{0.6775}
& \cellcolor{lightorange}\textbf{0.9555} 
& 0.6091 
& \textcolor{brown}{0.6316} 
& \textcolor{brown}{12}
\\
\hline

\multicolumn{19}{c}{\textit{32B open-source models}} \\ \hline
dsdistill-Qwen32B~\citep{ds-R1} 
& 0.6540 & 0.6696 & 0.3539 & \textcolor{brown}{0.5592}
& 0.3979 & 0.7358 & 0.7373 & \textcolor{brown}{0.6237}
& 0.2178 
& 0.4010 & 0.6697 & 0.7795 & 0.7229 & \textcolor{brown}{0.6433}
& 0.9225 
& 0.6177 
& \textcolor{brown}{0.6061}
& \textcolor{brown}{16}
\\

Qwen3-32B~\citep{qwen3} 
& 0.6247 & 0.6395 & 0.4255 & \textcolor{brown}{0.5632}
& 0.3806 & 0.7685 & 0.8038 & \textcolor{brown}{0.6510}
& 0.2016 
& 0.4355 & 0.7415 & 0.8206 & 0.7083 & \textcolor{brown}{0.6765}
& \textbf{0.9333} 
& 0.5025
& \textcolor{brown}{0.6143}
& \textcolor{brown}{15}
\\

QwQ-32B~\citep{qwen2.5} 
& 0.6791 & 0.6842 & 0.3905 & \textcolor{brown}{0.5846}
& 0.4085 & 0.7500 & 0.7296 & \textcolor{brown}{0.6294}
& 0.1749 
& 0.4305 & 0.6831 & 0.8470 & 0.7375 & \textcolor{brown}{0.6745}
& 0.9296
& 0.4642 
& \textcolor{brown}{0.6084}
& \textcolor{brown}{14}
\\

\hline
\multicolumn{19}{c}{\textit{14B open-source models}} \\ \hline
dsdistill-Qwen14B~\citep{ds-R1}
& 0.6635 & 0.5981 & 0.3404 & \textcolor{brown}{0.5340}
& 0.4169 & 0.7421 & 0.7750 & \textcolor{brown}{0.6447}
& 0.2470 
& 0.3823 & 0.6198 & 0.6660 & 0.7497 & \textcolor{brown}{0.6045}
& 0.8641
& 0.5795
& \textcolor{brown}{0.5880}
& \textcolor{brown}{16}
\\

Qwen3-14B~\citep{qwen3}
& 0.6540 & 0.6486 & 0.4820 & \textcolor{brown}{0.5949}
& 0.3871 & 0.7484 & 0.7540 & \textcolor{brown}{0.6298}
& 0.1680 
& 0.4215 & 0.7047 & 0.8102 & 0.6882 & \textcolor{brown}{0.6562}
& 0.8155 
& 0.6600
& \textcolor{brown}{0.6109}
& \textcolor{brown}{16}
\\

\hline
\multicolumn{19}{c}{\textit{7$\sim$8B open-source models}} \\ \hline
LLaMA3.1-8B~\citep{llama3} 
& 0.5871 & 0.5714 & 0.1957 & \textcolor{brown}{0.4514}
& 0.4476 & 0.7358 & 0.7581 & \textcolor{brown}{0.6472}
& 0.2147
& 0.2771 & 0.5791 & 0.7925 & 0.7096 & \textcolor{brown}{0.5896}
& 0.6805 
& 0.6703 
& \textcolor{brown}{0.5553}
& \textcolor{brown}{20}
\\

dsdistill-LLaMA-8B~\citep{ds-R1} 
& 0.5906 & 0.5507 & 0.2396 & \textcolor{brown}{0.4603}
& 0.4190 & 0.7294 & 0.7490 & \textcolor{brown}{0.6325}
& 0.2001 
& 0.3321 & 0.4430 & 0.6070 & 0.7587 & \textcolor{brown}{0.5352}
& 0.8157 
& 0.7617 
& \textcolor{brown}{0.5536}
& \textcolor{brown}{19}
\\

EmoLLM~\citep{EmoLLM} 
& 0.6180 & 0.5472 & 0.3322 & \textcolor{brown}{0.4991}
& 0.4334 & 0.7229 & 0.7931 & \textcolor{brown}{0.6498}
& 0.1408 
& 0.3373 & 0.4311 & 0.5749 & 0.6735 & \textcolor{brown}{0.5042}
& 0.7906 
& 0.6346 
& \textcolor{brown}{0.5407}
& \textcolor{brown}{21}
\\

Psyche-R1~\citep{Psyche-R1} 
& 0.5013 & 0.5222 & 0.2021 & \textcolor{brown}{0.4085}
& 0.4169 & 0.6877 & 0.7962 & \textcolor{brown}{0.6336}
& 0.2424 
& 0.5872 & 0.6989 & 0.8401 & 0.7194 & \textcolor{brown}{0.7114}
& 0.8164 
& 0.6954 
& \textcolor{brown}{0.5943}
& \textcolor{brown}{16}
\\

Qwen3-8B~\citep{qwen3} 
& 0.5941 & 0.6395 & 0.4057 & \textcolor{brown}{0.5464}
& 0.4169 & 0.7358 & 0.7515 & \textcolor{brown}{0.6347}
& 0.1489 
& 0.4009 & 0.5824 & 0.7927 & 0.6899 & \textcolor{brown}{0.6165}
& 0.8115 
& 0.4773 
& \textcolor{brown}{0.5729} 
& \textcolor{brown}{20}
\\

\hline
Mindora$_{SFT}$ 
& 0.5693 & 0.6207 & 0.4472 & \textcolor{brown}{0.5457}
& 0.4887 & 0.7437 & 0.7597 & \textcolor{brown}{0.6640}
& {0.4193} 
& 0.5830 & 0.7263 & 0.8535 & 0.7376 & \textcolor{brown}{0.7251}
& 0.7764 
& 0.5512
& \textcolor{brown}{0.6367}
& \textcolor{brown}{13}
\\
Mindora$_{RL}$ 
& 0.6112 & 0.6301 & 0.4279 & \textcolor{brown}{0.5564}
& 0.3849 & 0.7358  & 0.7140  & \textcolor{brown}{0.6116}
& 0.1456
& 0.5844 & 0.8020 & \cellcolor{lightorange}\textbf{0.9001} & 0.7808  & \textcolor{brown}{0.7668}
&  0.8015
&  0.4493
& \textcolor{brown}{0.6129}
& \textcolor{brown}{16}
\\
Mindora$_{SFT+RL}$ 
& 0.5975 & 0.6207 & 0.4566 & \textcolor{brown}{0.5583}
& \cellcolor{lightorange}\textbf{0.5240} & 0.7715 & \cellcolor{lightorange}\textbf{0.8140} & \cellcolor{lightorange}\textcolor{brown}{\textbf{0.7032}}
& 0.3803
& 0.5996 & 0.7839 & 0.8593 & 0.8212 & \textcolor{brown}{0.7660}
& 0.7159
& 0.5681 
& \textcolor{brown}{0.6548} 
& \textcolor{brown}{10}
\\
Mindora w/o S1
& 0.6912 & 0.6456 & \cellcolor{lightorange}\textbf{0.5262} & \textcolor{brown}{0.6210}
& 0.4252 & 0.7013 & 0.8106 & \textcolor{brown}{0.6457}
& \cellcolor{lightorange}\textbf{0.4719}
& 0.5927 & 0.7716 & 0.8737 & 0.8133 & \textcolor{brown}{0.7628}
&  0.7982
& 0.6286
& \textcolor{brown}{0.6731}
& \textcolor{brown}{8}
\\
Mindora w/o S2
& 0.6044  & 0.6270  & 0.5240 & \textcolor{brown}{0.5851}
& 0.4576 & 0.7730  & 0.8225  & \textcolor{brown}{0.6844}
&  0.4430
& 0.5486  & 0.7330  & 0.8416  & 0.6308  & \textcolor{brown}{ 0.6885 }
&  0.7955
&  0.6038
& \textcolor{brown}{0.6465 }
& \textcolor{brown}{12}
\\
Mindora w/o S3
& 0.6729 & 0.6637 & {0.5240} & \textcolor{brown}{0.6202}
& 0.4868 & 0.7685 & 0.7806 & \textcolor{brown}{0.6786}
& {0.4400}
& 0.6145 & 0.7778 & 0.8695 & \cellcolor{lightorange}\textbf{0.8516} & \cellcolor{lightorange}\textcolor{brown}{\textbf{0.7784}}
& 0.6948
& 0.6796
& \textcolor{brown}{0.6788}
& \textcolor{brown}{7}
\\
Mindora 
& 0.7293 & 0.6842 & {0.5088} & \cellcolor{lightorange}\textcolor{brown}{\textbf{0.6408}}
& 0.4655 & \cellcolor{lightorange}\textbf{0.7760} & 0.8030 & \textcolor{brown}{0.6815}
& 0.4016 
& \cellcolor{lightorange}\textbf{0.6317} & 0.7590 & 0.8535 & {0.8442} & \textcolor{brown}{0.7721}
& 0.8379 
& 0.7178 
& \cellcolor{lightorange}\textcolor{brown}{\textbf{0.6933}} 
& \cellcolor{lightorange}\textcolor{brown}{\textbf{5}} 
\\
\hline
\end{tabular}
  }
\end{table*}

\begin{table*}[t]
\centering
\renewcommand{\arraystretch}{1.3}
\caption{Multi-dimensional evaluation of reasoning-trajectory quality across models.}
\label{tab:trajectory_main}
\resizebox{0.8\linewidth}{!}{
\begin{tabular}{lcccccccc}
\hline
\textbf{Model} 
& \textbf{Accuracy}
& \makecell[c]{\textbf{Terminology}\\\textbf{Parsimony $\uparrow$ }}
& \makecell[c]{\textbf{Step}\\\textbf{Adequacy $\uparrow$}}
& \makecell[c]{\textbf{Avg Steps$\downarrow$}}
& \makecell[c]{\textbf{Logical}\\\textbf{Coherence $\uparrow$}}
% & \textbf{Leap Rate} $\downarrow$
& \makecell[c]{\textbf{Irrelevance}\\\textbf{Rate $\downarrow$}}
& \makecell[c]{\textbf{Contradiction}\\\textbf{Rate $\downarrow$}}
& \textbf{Informativeness} $\uparrow$ \\
\hline
GPT-o4-mini 
& 0.4593
& 0.8750 
& \textbf{0.9042} & 5.1681
& 0.8875 
% & 0.0490 
& 0.0064 & 0.0101 & 0.7500 \\
Deepseek-R1
& 0.5102
& 0.7458 
& 0.6750 & 7.7480
& 1.0542 
% & \textbf{0.0184} 
& 0.0138 & 0.0410 & 0.9958 \\
Qwen-plus 
& 0.4555
& 0.6208 
& 0.4500 & 9.2035
& 0.8708
% & 0.0379 
& 0.0306 & 0.0263 & 0.9042 \\
LLaMA-4 
& 0.4882
& \textbf{0.9375} 
& 0.9333 & 5.0826
& 0.9292 
% & 0.0302 
& 0.0092 & \textbf{0.0049} & 0.7917 \\
Psyche-R1 
& 0.4396
& 0.7250 
& 0.6458 & 5.6422
& 0.8708 
% & 0.0516
& 0.0074 & 0.0233 & 0.7667 \\
Qwen3-8B 
& 0.4614
& 0.6792 
& 0.4208 & 7.8696
& 0.8625
% & 0.0333 
& 0.0286 & 0.0214 & 0.9083 \\ \hline
Mindora  w/o S1 
& 0.4969
& 0.8458 
& 0.6958 & 6.2377
& 0.9625 
% & 0.0319 
& \textbf{0.0060} & 0.0163 & 0.8875 \\
Mindora  w/o S2
& 0.5223
& 0.5542 
& 0.1917 & 10.0391
& 0.9917 
% & 0.0553
& 0.0160 & 0.0383 & \textbf{1.0042} \\
Mindora  w/o S3
& 0.5199
& 0.8750 
& 0.8250 & 5.2248
& 1.0042 
% & 0.0393
& 0.0102 & 0.0115 & 0.9250 \\
Mindora  
& \textbf{0.5467}
& 0.9042
& 0.8625 & \textbf{5.0444}
& \textbf{1.0625} 
% & 0.0488 
& 0.0065 & 0.0154 & 0.9917 \\
\hline
\end{tabular}}
\end{table*}

\subsection{Experimental Setup}
\label{sec:experimental_setup}

We evaluate 20 LLMs on \BENCH{} to examine three questions:
(i) whether strong final-answer competence is accompanied by evidence-constrained judgment;
(ii) whether restraint failures recur across task families that transform different forms of evidence into different types of judgment; and
(iii) whether post-training targeted at evidence-to-judgment transitions produces consistent improvements across heterogeneous tasks.
To address the first two questions, we evaluate proprietary general-purpose and reasoning models, large open-weight models, compact open-weight models, and mental-health-oriented baselines. We examine the third question through \MODEL{} and its ablation variants.

\paragraph{Models.}
We evaluate a broad set of LLMs, including proprietary systems, open-weight models across multiple scales, reasoning-oriented variants, and mental-health-specific baselines. This grouping allows us to compare the effects of model scale, general reasoning post-training, and domain-specific mental-health adaptation under the same benchmark. Model details are provided in Appendix~\ref{append:baseline_models}.

\paragraph{Evaluation protocol.}
All general-purpose models except Qwen3-8B are accessed through official APIs, while domain-specific models and Qwen3-8B are deployed on a single NVIDIA A800-SXM4-80GB GPU. We keep decoding parameters at their default values unless otherwise specified. All models are prompted to produce both a final answer and a structured reasoning trajectory, enabling evaluation of task accuracy and trajectory reliability under a unified protocol. To examine whether the structured prompting format itself explains the observed gains, we additionally conduct a few-shot ICL analysis in Appendix~\ref{append:icl_analysis}.

\subsection{Main Benchmark Results}

Table~\ref{tab:main_result} reports final-answer performance across the six task families. Because the tasks use different metrics, we use average rank to summarize cross-task performance. Clear rank reversals emerge across task families. GPT-o4-mini and DeepSeek-R1 perform strongly on psychiatry QA, with average scores of 0.7402 and 0.7459, but obtain only 0.2086 and 0.2020 on intervention. Similarly, Qwen2.5-72B scores 0.6527 on condition detection and 0.9555 on abstraction, but only 0.2796 on intervention. Thus, competence on structured knowledge tasks does not consistently transfer to narrative interpretation or therapeutic-action selection.

\MODEL{} achieves the best average rank and improves over Qwen3-8B across all six task families, indicating that its gains are not confined to one dataset or output format. To further examine whether these gains are confined to \BENCH{}, we evaluate \MODEL{} and its Qwen3-8B backbone on 1,012 test instances from GoEmotions, an external fine-grained emotion classification dataset (Appendix~\ref{append:external_evaluation}). \MODEL{} achieves 0.4407 accuracy, compared with 0.3528 for Qwen3-8B, which provides complementary evidence beyond \BENCH{}. 

\subsection{Key Findings of \BENCH{}}
\label{sec:key_findings}

\paragraph{Finding 1: Answer competence does not imply evidence-constrained judgment.}
Strong endpoint performance is not consistently accompanied by well-controlled reasoning. DeepSeek-R1 achieves high Informativeness (0.9958), but produces 7.748 steps on average and has a Contradiction Rate of 0.0410. Qwen-plus is similarly informative (0.9042), yet shows low Step Adequacy (0.4500), long trajectories (9.2035 steps), and relatively high irrelevance (0.0306). A representative case further shows that models reaching the same correct label may rely on under-specified templates or exhaustive option scanning rather than a concise evidence-to-label mapping (Appendix~\ref{append:general_case_study}). Thus, final-answer correctness can conceal substantial differences in reasoning selectivity, grounding, and internal stability.

\paragraph{Finding 2: The restraint gap recurs across evidence-to-judgment settings.}
Models strong on psychiatry QA or misinformation verification do not consistently dominate appraisal, condition detection, or intervention. Although these tasks differ in evidence form and output type, they expose related failures: overinterpreting sparse cues, selecting overly specific actions, introducing unsupported clinical assumptions, or forming claims stronger than the evidence warrants. The restraint gap therefore recurs across task families rather than being confined to a single dataset or output format.

\paragraph{Finding 3: Targeted trajectory training supports transferable evidence-to-judgment control.}
\MODEL{} achieves the best average rank (4.92) and improves over Qwen3-8B across all six task families. Removing hard-case filtering worsens the average rank to 8.46 and increases trajectory length; removing structured rewriting causes the largest degradation (11.62), together with lower Step Adequacy and more contradiction; and removing the consistency reward worsens the rank to 7.08. Representative cases respectively show unnecessary speculation beyond observed evidence, backtracking between competing judgments, and the introduction of unsupported clinical criteria (Appendix~\ref{append:ablation_details}). These behavior-level patterns support the interpretation that controlling evidence-to-judgment transitions transfers across heterogeneous tasks, without establishing latent reasoning faithfulness or clinical validity.

\subsection{Reasoning-Trajectory Evaluation}
\label{sec:trajectory_results}

We evaluate trajectory quality on a 5\% sample using the protocol in Section~\ref{sec:trajectory_eval}; trajectories associated with incorrect final answers receive zero scores so that the primary evaluation reflects both endpoint correctness and the quality of the rationale supporting it. As shown in Table~\ref{tab:trajectory_main}, \MODEL{} achieves high Logical Coherence (1.0625) and Informativeness (0.9917), while maintaining low Irrelevance and Contradiction Rates (0.0065 and 0.0154). Compared with Qwen3-8B, it improves Terminology Parsimony from 0.6792 to 0.9042 and Step Adequacy from 0.4208 to 0.8625, while reducing the average trajectory length from 7.8696 to 5.0444. Compared with DeepSeek-R1, \MODEL{} retains similar Informativeness but substantially lowers the Contradiction Rate from 0.0410 to 0.0154. These results show that similarly informative trajectories can differ substantially in conciseness, adequacy, and internal consistency, revealing differences in how models control necessary, evidence-supported, and judgment-aligned inferences that final-answer accuracy alone cannot capture.

Conditioning the evaluation on correct final predictions yields the same pattern (Appendix~\ref{append:correct_conditioned_trajectory}). \MODEL{} produces shorter trajectories than Qwen3-8B, with higher Step Adequacy and lower Irrelevance Rate. Removing structured rewriting again worsens trajectory length, adequacy, irrelevance, and contradiction. The improvements are therefore not explained solely by final-answer accuracy.

\section{Conclusion}

We introduced \BENCH{}, a unified benchmark that formulates reliable mental-health reasoning as evidence-constrained judgment: whether the specificity, certainty, severity, and actionability of a model's interpretations, decisions, and claims remain proportional to the available evidence. Across six task families and 13 datasets, our evaluation of 20 LLMs reveals a recurring restraint gap, in which models may overinterpret sparse cues, introduce unsupported clinical assumptions, or produce rationales that do not justify the strength of their conclusions. \BENCH{} audits these observable evidence-to-judgment failures through trajectory-level measures of usability, logical structure, and informational contribution, revealing differences that final-answer correctness alone cannot capture. As an intervention test of this benchmark hypothesis, we train \MODEL{}, an 8B model that achieves the best average rank and improves over its backbone across all six task families. Its cross-task gains and ablation results support the interpretation that restraint over evidence-to-judgment transitions can be targeted through hard-case supervision, structured trajectories, and consistency-aware optimization. These results position \BENCH{} as both an evaluation framework and a research substrate for evidence-constrained mental-health reasoning, while real-world deployment remains outside its scope and requires clinical validation, safety monitoring, and human oversight.

\section*{Limitations}
\BENCH{} provides a controlled and reproducible testbed for reliable mental-health reasoning, but it does not cover the full scope of mental-health practice. The current benchmark focuses on single-instance reasoning over posts, counseling dialogues, psychiatry questions, systematic-review abstracts, and online claims, while longitudinal therapy, multi-session risk monitoring, clinician--patient collaboration, culturally situated care, and multimodal behavioral signals remain future directions. In addition, \BENCH{} combines newly curated resources with task-adapted existing datasets, so some tasks inherit source-specific assumptions, such as weak supervision in social-media condition detection or strategy-level abstraction in counseling intervention. Our reasoning-trajectory protocol operationalizes usability, logical structure, and informational contribution and is validated against expert annotations, but broader expert review and cross-cultural validation would further strengthen the benchmark. 

\section*{Ethical Statement}
\MODEL{} is used only as a benchmark-guided post-training case study. Strong benchmark performance does not imply clinical safety, treatment validity, deployment readiness, or regulatory compliance; real-world mental-health applications require clinical validation, human oversight, safety monitoring, and safeguards against unsupported advice, over-pathologizing, and crisis mismanagement.

\section*{Acknowledgments}
This work is supported by the Key Project of the National Natural Science Foundation of China (U23A20316), CCF-Tencent Rhino-Bird Open Research Fund (CCF-Tencent RAGR20250115), Wuhan Natural Science Foundation Exploratory Program (Morning Light Program) Project (2026040301020029) and the Intelligent Computing Center of the National Cybersecurity Talent and Innovation Base, Wuhan

% Custom bibliography entries only
\bibliography{custom}

\clearpage
\appendix
\section{Related Work}

\subsection{Psychiatric Diagnosis and Clinical Decision-Making Benchmarks}
Recent benchmarks have advanced the evaluation of LLMs from social-media mental-health classification toward clinically grounded psychiatric reasoning. MentalBench evaluates DSM-5-grounded diagnostic decision-making using a psychiatrist-built knowledge graph and controlled synthetic cases with varying information completeness and diagnostic complexity~\citep{song2026mentalbench}. MentalDx Bench targets real-world disorder-level diagnosis from de-identified EHRs annotated under ICD-11 guidelines, while MentalSeek-Dx further models progressive hypothetico-deductive diagnostic reasoning~\citep{sun2026mentalseekdx}. LingxiDiagBench extends this direction to Chinese psychiatric consultation by evaluating both static diagnostic inference and dynamic multi-turn diagnosis over EMR-aligned synthetic dialogues~\citep{xu2026lingxidiagbench}. Beyond diagnosis, PsychiatryBench covers multiple psychiatry-focused QA tasks, including diagnosis, treatment planning, longitudinal follow-up, management planning, and sequential case analysis~\citep{fouda2026psychiatrybench}, while MENTAT evaluates real-world ambiguity across diagnosis, treatment, monitoring, triage, and documentation using clinician-annotated decision scenarios~\citep{lamparth2025moving}.

These benchmarks provide important clinical grounding, but they mainly evaluate LLMs as psychiatrists or mental-health practitioners making clinical decisions. In contrast, \textsc{MiraMind} treats mental-health reasoning as a broader set of cognitive operations, including appraisal, diagnosis, intervention, abstraction, and verification. This broader scope reflects user-facing mental-health interactions, where models must often interpret ambiguous narratives, identify risk without over-pathologizing distress, select helpful responses, summarize clinically relevant content, and assess the credibility of mental-health advice before any externally verifiable outcome is available.

\subsection{Counseling and Psychotherapy Competence Benchmarks}
Recent benchmarks also evaluate LLMs in counseling and psychotherapy-oriented settings. CBT-Bench assesses cognitive behavioral therapy assistance across basic CBT knowledge, cognitive model understanding, and therapeutic response generation~\citep{zhang2025cbtbench}. CounselingBench evaluates 22 general-purpose and medical-finetuned LLMs on five core mental-health counseling competencies using NCMHCE-derived questions~\citep{nguyen2025counselingbench}. CounselBench provides a large-scale expert evaluation and adversarial benchmark for single-turn mental-health counseling, with 100 mental-health professionals rating model and human responses and adversarial questions designed to expose counseling-specific failures~\citep{li2025counselbench}. MindEval further evaluates LLMs in realistic multi-turn mental-health support conversations using simulated patients and automatic evaluation validated against licensed clinical psychologists~\citep{pombal2025mindeval}.

While these benchmarks are clinically valuable, they primarily evaluate counseling behavior or therapeutic response quality. \textsc{MiraMind} addresses a different evaluation target: whether LLMs can perform reliable mental-health reasoning across heterogeneous task families. Rather than focusing only on whether a model behaves like a counselor or therapist, \textsc{MiraMind} evaluates clinically aligned reasoning over appraisal, diagnosis, intervention, abstraction, and verification, together with reasoning qualities such as grounding, coherence, conciseness, hallucination avoidance, and internal consistency. In this sense, \textsc{MiraMind} complements psychotherapy competence benchmarks by testing the broader reasoning operations that underlie mental-health interpretation and assessment beyond therapeutic dialogue alone.

\subsection{Safety, Crisis, and Ethical Reasoning Benchmarks}
Safety-oriented benchmarks evaluate whether LLMs can handle high-risk mental-health scenarios. The Mental Health Crisis Benchmark evaluates crisis classification and response quality across categories such as suicidal ideation, self-harm, anxiety crisis, substance abuse or withdrawal, risk-taking behavior, and violent thoughts~\citep{arnaiz2026between}. EthicsMH targets ethical reasoning in therapeutic and psychiatric contexts, with scenarios involving confidentiality, autonomy, beneficence, fairness, and bias, together with structured decision options, expert-aligned reasoning, expected behavior, real-world impact, and stakeholder perspectives~\citep{kasu2025ethicsmh}. CounselBench-Adv similarly stress-tests LLMs with expert-authored adversarial counseling questions designed to elicit counseling-specific failures~\citep{li2025counselbench}.

These benchmarks are indispensable for auditing high-risk failures, but they usually target specific failure domains such as crisis response, ethical dilemmas, or adversarial counseling. \textsc{MiraMind} complements this line of work by evaluating general mental-health reasoning reliability across both ordinary and high-stakes tasks. Rather than focusing only on whether a model responds safely to predefined risk scenarios, \textsc{MiraMind} evaluates whether it can reason coherently, cautiously, and consistently across appraisal, diagnosis, intervention, abstraction, and verification.

\clearpage
\section{\textsc{MiraMind} Benchmark Details}
\label{append:benchmark_details}

\subsection{Dataset Licenses and Access Conditions}
\label{append:dataset_licenses}

Table~\ref{tab:dataset_licenses} summarizes the licenses and access conditions of the datasets used in \BENCH{}. For datasets without an explicit public license, we follow the original dataset access terms and cite the corresponding source. For restricted or platform-derived datasets, we do not redistribute raw texts and release only processed metadata, prompts, labels, or evaluation scripts where permitted.

\begin{table*}[t]
\centering
\renewcommand{\arraystretch}{1.2}
\caption{Dataset licenses and access conditions for \BENCH{}. ``Not specified'' indicates that we did not find an explicit dataset-specific license in the original release page or paper. For restricted or platform-derived datasets, we follow the original access terms and do not redistribute raw texts unless explicitly permitted.}
\label{tab:dataset_licenses}
\resizebox{\linewidth}{!}{
\begin{tabular}{llll}
\hline
\textbf{Dataset} 
& \textbf{Original Source / Release}
& \textbf{License / Access Condition}
& \textbf{Redistribution Note}
\\
\hline

CognitiveReframing
& \makecell[l]{Original GitHub release\\ \citep{sharma2023cognitive}}
& \makecell[l]{CC BY-NC-ND 4.0}
& \makecell[l]{Non-commercial research use;\\ no redistribution of modified data.}
\\ 
\addlinespace[5pt]

PatternReframe
& \makecell[l]{ParlAI / original release\\ \citep{Mounica2023Training}}
& \makecell[l]{Released through ParlAI;\\ dataset-specific license not clearly specified}
& \makecell[l]{Use follows original release terms;\\ processed prompts/scripts only where permitted.}
\\ \addlinespace[5pt]

Therapist Q\&A
& \makecell[l]{Kaggle Therapist Q\&A corpus\\ \citep{shreevastava2021detecting}}
& \makecell[l]{Kaggle-hosted dataset;\\ license follows original Kaggle page}
& \makecell[l]{Use follows original dataset terms;\\ raw redistribution not assumed.}
\\ \addlinespace[5pt]

DepSign
& \makecell[l]{LT-EDI shared-task dataset\\ \citep{depsign}}
& \makecell[l]{Shared-task access;\\ explicit public license not specified}
& \makecell[l]{Use follows shared-task terms;\\ raw redistribution not assumed.}
\\ \addlinespace[5pt]

SWMH
& \makecell[l]{Original release / Zenodo\\ \citep{ji2022suicidal}}
& \makecell[l]{Restricted research-only access\\ under data access agreement}
& \makecell[l]{Redistribution not permitted;\\ users should request access from source.}
\\ \addlinespace[5pt]

T-SID
& \makecell[l]{Original Twitter-derived dataset\\ \citep{ji2022suicidal}}
& \makecell[l]{Twitter-derived data;\\ explicit dataset license not specified}
& \makecell[l]{Use follows original dataset and platform terms;\\ raw tweet redistribution not assumed.}
\\ \addlinespace[5pt]

PsyDTCorpus$_M$
& \makecell[l]{Processed from PsyDTCorpus\\ \citep{xie2025psydt}}
& \makecell[l]{Released by original authors;\\ license follows original release}
& \makecell[l]{Used only for training;\\ processed version follows original terms.}
\\ \addlinespace[5pt]

AnnoMI$_M$
& \makecell[l]{Processed from AnnoMI\\ \citep{annomi2022}}
& \makecell[l]{Public dataset release;\\ dataset-specific license not clearly specified}
& \makecell[l]{Use follows original access terms;\\ processed labels/prompts released where permitted.}
\\ \addlinespace[5pt]

MHQA
& \makecell[l]{Original MHQA release\\ \citep{racha2025mhqa}}
& \makecell[l]{Dataset-specific license to be confirmed\\ from original release}
& \makecell[l]{Use follows original release terms;\\ raw redistribution not assumed.}
\\ \addlinespace[5pt]

MedQA$_M$
& \makecell[l]{Processed from MedQA\\ \citep{jin2020disease}}
& \makecell[l]{Official dataset license not clearly specified}
& \makecell[l]{Psychiatry subset extracted;\\ redistribution follows original MedQA terms.}
\\ \addlinespace[5pt]

MedMCQA$_M$
& \makecell[l]{Processed from MedMCQA\\ \citep{pal2022medmcqa}}
& \makecell[l]{Apache-2.0}
& \makecell[l]{Psychiatry subset extracted;\\ redistribution permitted under Apache-2.0 terms.}
\\ \addlinespace[5pt]

PubMedQA$_M$
& \makecell[l]{Processed from PubMedQA\\ \citep{jin2019pubmedqa}}
& \makecell[l]{MIT License}
& \makecell[l]{Psychiatry subset extracted;\\ redistribution permitted under MIT terms.}
\\ \addlinespace[5pt]

PSRS
& \makecell[l]{Newly curated from Cochrane\\ Library abstracts}
& \makecell[l]{Cochrane/Wiley terms;\\ full abstract text may be copyright restricted}
& \makecell[l]{We release review identifiers, prompts,\\ expert scores, and annotations where permitted;\\ full abstract text is not redistributed unless allowed.}
\\ \addlinespace[5pt]

MentalMisinfo
& \makecell[l]{Original MentalMisinfo release\\ \citep{nguyen2025supporters}}
& \makecell[l]{Released by original authors;\\ explicit dataset license not clearly specified}
& \makecell[l]{Use follows original dataset terms;\\ raw video/transcript redistribution not assumed.}
\\

\hline
\end{tabular}
}
\end{table*}

\subsection{Usage Declaration}
\BENCH{} is intended for academic research on evaluating and analyzing mental-health reasoning in LLMs. We use existing datasets only for benchmark construction, model evaluation, and benchmark-guided post-training within research contexts. For datasets with explicit licenses or access conditions, we follow the original terms. For restricted, platform-derived, or license-unspecified resources, we do not assume permission to redistribute raw data and release only derived prompts, metadata, labels, evaluation scripts, or annotations where permitted. Processed datasets marked with $_M$ are intended for research evaluation and training under the access conditions of their original sources. PSRS is intended for evaluating evidence-grounded psychiatry summarization; for Cochrane-derived materials, we release review identifiers, prompts, expert annotations, and scoring metadata where permitted, but do not redistribute full copyrighted abstract text unless allowed by the original terms. Licenses of datasets are detailed in Appendix~\ref{append:dataset_licenses}.

\subsection{Personally Identifying Information and Sensitive Content}
\label{append:pii_sensitive_content}

\BENCH{} includes mental-health-related posts, counseling dialogues, online claims, and psychiatry questions, which may contain sensitive personal experiences or stigmatizing language. We therefore treat all user-generated and dialogue-based data as potentially sensitive. For existing datasets, we rely on the anonymization and access-control procedures of the original releases and follow their redistribution terms. For processed datasets marked with $_M$, we do not introduce new personal identifiers; client utterances are compressed into task-focused case summaries, and outputs are standardized into labels, prompts, or evaluation items. For restricted, platform-derived, or license-unspecified resources, we do not redistribute raw text unless explicitly permitted by the original source.

For newly curated PSRS, the source materials are publicly available systematic-review abstracts rather than patient-level records, so no individual patient identifiers are included. For all released artifacts, we will avoid publishing personally identifying information and will release only permitted metadata, prompts, labels, evaluation scripts, expert scores, and annotations where allowed. Because mental-health data can include distressing, stigmatizing, or crisis-related content, benchmark users should treat the data as sensitive research material and use it only under appropriate ethical and institutional safeguards.

\subsection{Model Size and Computational Budget}
\label{append:compute_budget}

For benchmark evaluation, most proprietary, hosted, and large open-weight models are accessed through official APIs, including GPT-o4-mini, GPT-4o, DeepSeek-R1, DeepSeek-V3, Qwen-plus, QwQ-plus, LLaMA-4, LLaMA-3.3-70B, DeepSeek-distilled 70B/32B/14B variants, Qwen2.5-72B, Qwen3-32B/14B, and QwQ-32B. Their exact provider-side compute budgets are not publicly available. The locally evaluated models include EmoLLM, Psyche-R1, and Qwen3-8B, all at the 7--8B scale. Local benchmark inference was conducted on 8 NVIDIA GPUs for approximately 2 hours in total, corresponding to about 16 GPU-hours.

Reasoning-trajectory evaluation is performed with gpt-5-mini-2025-08-07 as an API-based LLM-as-Judge. Combining answer generation through APIs and trajectory evaluation, the total monetary cost for running the benchmark experiments was approximately \$20 under the API pricing used in our experiments. \MODEL{} is initialized from Qwen3-8B and therefore has approximately 8B parameters; its benchmark inference is included in the local compute budget above.

\subsection{Appraisal Dataset Details}
\label{append:appraisal_dataset_details}
Detailed descriptions of these datasets are shown below:

\begin{itemize}[leftmargin=1em]
\item CognitiveReframing~\citep{sharma2023cognitive} combines simulated negative thoughts from the Thought Records Dataset~\citep{burger2021natural} and self-reports from the Mental Health America website\footnote{\url{https://screening.mhanational.org/}},
 with distortions annotated by 15 trained mental-health professionals.
\item PatternReframe~\citep{Mounica2023Training} is constructed from PERSONA-CHAT personas~\citep{zhang2018personalizing} and contains statements crafted to manifest specific distortions and labeled by five independent raters. 
\item Therapist Q\&A~\citep{shreevastava2021detecting}, derived from real therapist–client interactions in the Therapist Q\&A corpus\footnote{\url{https://www.kaggle.com/arnmaud/therapist-qa}},
 provides naturally occurring distorted statements annotated by two clinical raters.
\end{itemize}

These three datasets adopt slightly different taxonomies of cognitive errors.
\begin{itemize}[leftmargin=1em]
\item Shared across all three datasets:
All-or-Nothing Thinking; Overgeneralization; Labeling; Fortune Telling; Mind Reading; Should Statements; Personalization.
\item Unique to CognitiveReframing: Emotional Reasoning; Comparing and Despairing; Blaming; Negative Feeling or Emotion; Catastrophizing; Discounting the Positive.
\item Unique to PatternReframe: Mental Filtering; Catastrophizing; Discounting the Positive.
\item Unique to Therapist Q\&A: Emotional Reasoning; Mental Filtering; Magnification; No Distortion.
\end{itemize}

\subsection{Intervention Dataset Details}
\label{append:intervention_dataset_detail}
We instantiate this dimension through the counseling-strategy formulation task, covering thirteen commonly taught intervention types. The strategies include Clarification, Paraphrasing, Reflection of Feeling, Summarizing, Questioning Skills, Immediacy, Use of Silence, Self-Disclosure, Confrontation, Encouragement, Repetition, Interpretation, and Guidance.

To construct evaluation data, we use two high-quality counseling dialogue corpora. Client utterances are compressed into concise case summaries using GPT-4o prompts, and counselor utterances are annotated to extract strategy labels as reference answers. All constructed items are manually reviewed and verified by an expert counselor with over ten years of clinical experience, resulting in the final $[\text{dataset}]_M$ versions. 
We deliberately separate the training and test sources for this task: PsyDTCorpus$_M$ is used only for training, while AnnoMI$_M$ is used for evaluation. This design tests whether intervention-planning ability transfers from the partially synthetic training corpus to real counseling dialogue demonstrations.

\begin{itemize}[leftmargin=1em]
\item {PsyDTCorpus$_M$}: Based on PsyDTCorpus~\citep{xie2025psydt}, which contains 5,000 high-quality single-turn dialogues from SoulChatCorpus and 12 anonymized real counseling cases synthesized into multi-turn interactions. Due to its partially synthetic nature, this dataset is used to train Mindora but is excluded from benchmark evaluation. 
\item {AnnoMI$_M$}: Based on AnnoMI~\citep{annomi2022}, constructed from authorized motivational interviewing (MI) demonstration videos sourced from YouTube and Vimeo, transcribed and curated into high-quality multi-turn counseling interactions.\end{itemize}

\subsection{Multi-step Dataset Details}
\label{append:qa_raw_data_info}
We instantiate this dimension through a psychiatry QA task and evaluate it using four psychiatry-focused datasets.

\begin{itemize}[leftmargin=1em]
\item MHQA~\citep{racha2025mhqa} provides knowledge-intensive question-answer pairs from 471k PubMed abstracts covering major mental-health conditions. 
\item MedQA$_{en}$~\citep{jin2020disease} provides exam-style multiple-choice medical questions with clinical vignettes collected from professional medical licensing resources, covering broad disease knowledge and diagnostic decision-making.
\item MedMCQA~\citep{pal2022medmcqa} provides large-scale multiple-choice questions curated from Indian medical entrance and licensing examinations, spanning diverse medical specialties and knowledge-intensive clinical reasoning.
\item PubMedQA~\citep{jin2019pubmedqa} provides research-focused question-answer pairs grounded in PubMed abstracts, requiring evidence-based inference over biomedical studies with yes/no/maybe-style conclusions.
\end{itemize}

\subsection{PSRS Dataset Construction}
\label{append:abstraction_dataset_construction}
In this section, we illustrate the construction pipeline of {PSRS}, a psychiatry-oriented systematic-review summarization resource derived from structured Cochrane abstracts and paired with expert evaluation of LLM-generated summaries. This dataset is introduced to fill a gap not covered by existing mental-health benchmarks: evaluating whether LLMs can abstract clinically relevant conclusions from psychiatric systematic-review evidence, rather than only answer diagnostic or counseling questions. The overall design follows a {source-abstract $\rightarrow$ structured-summary generation $\rightarrow$ expert rubric-based assessment workflow. 

\subsubsection{Raw Data Collection}
\label{app:psrs_source}
We curate 115 structured abstracts of systematic reviews from the Cochrane Library, restricted to the mental-health domain to ensure clinically coherent populations, interventions, and outcomes. Abstracts are sampled to preserve diversity in (i) target conditions, (ii) intervention families, and (iii) population characteristics. We retain review identifiers for traceability and use only publicly available abstract text, without any patient-level or trial-level records.

\subsubsection{LLM Summary Generation}
\label{app:psrs_generation}
For each Cochrane abstract, we generate four structured summaries using GPT-4o, GPT-4o-mini, Claude Opus, and LLaMA-3.1-70B-Instruct, yielding 460 summaries in total. To minimize format variance, we enforce a fixed schema and focus evaluation on a single target field, Main Results, which summarizes the core outcome-level findings supported by the abstract evidence. We use low-temperature decoding (temperature $\le 0.3$) for determinism and provide no external content beyond the abstract.

\subsubsection{Human Annotation}
\label{app:psrs_annotation_setup}
\paragraph{Annotators.}
Two clinical domain experts (psychiatrists) conduct the evaluation. Each generated \textit{Main Results} statement is assessed under a structured rubric with mandatory rationales for non-maximal labels, enabling auditing and error analysis.

\paragraph{Pilot calibration.}
Before the main annotation, we run a pilot on 10 abstracts per expert to refine label definitions and align judgment thresholds (e.g., borderline cases involving partial reporting vs.\ over-interpretation).

\subsubsection{Evaluation Rubric and Labeling Scheme}
\label{app:psrs_rubric}
Since our main paper evaluates only the Main Results field, we adopt a single-dimension scoring rubric.

\paragraph{Main Results (0--2).}
Annotators score the Main Results content on an ordinal scale:
\begin{itemize}[leftmargin=1em]
  \item 2 = accurate and complete with respect to the Cochrane abstract;
  \item 1 = partially correct / incomplete / contains minor inaccuracies;
  \item 0 = missing, unsupported, or substantially incorrect.
\end{itemize}
Rationales are mandatory for scores in $\{0,1\}$. In addition, annotators record a corrective action indicating the minimal edit needed to align the content with the abstract: add, remove, modify, rewrite, or none.

\subsubsection{Inter-Annotator Agreement and Quality Assurance}
\label{app:psrs_iaa}
Because full double-annotation is expensive with clinical experts, we adopt a targeted QA protocol.

\paragraph{Cross-check subset.}
After the main evaluation, we randomly select 30 reviews per model (i.e., 120 summaries in total) for cross-checking. In this phase, each expert reviews the other's prior Main Results labels and records agreement vs.\ disagreement.

\paragraph{Agreement reporting (IAA proxy).}
Rather than reporting chance-corrected coefficients (e.g., Cohen's $\kappa$), we report absolute disagreement counts on the QA subset. The IAA results are shown in Table~\ref{tab:psrs_iaa}. Notes that we report absolute disagreement counts due to the modest QA sample size. Disagreements mainly arise from borderline cases where summaries partially report effect directions/magnitudes or use stronger-than-supported wording, requiring subjective thresholds even among experts.

\begin{table}[h]
\centering
\small
\renewcommand{\arraystretch}{1.15}
\setlength{\tabcolsep}{6pt}
\caption{Inter-annotator agreement (IAA) on the QA subset, reported as the number of disagreements between the two psychiatrists for the Main Results dimension ($n=30$ reviews per model; $120$ summaries in total).}
\label{tab:psrs_iaa}
\resizebox{\linewidth}{!}{
\begin{tabular}{lcccc}
\hline
Model & GPT-4o & Claude Opus & LLaMA-3.1-70B & GPT-4o-mini \\
\hline
Main Results & 4 & 3 & 5 & 4 \\
\hline
\end{tabular}
}
\end{table}

\subsubsection{Output Artifacts and Release}
\label{app:psrs_release}
For each generated summary, PSRS releases (i) the structured Main Results output, (ii) expert scores, (iii) rationales for non-maximal judgments, and (iv) QA cross-check agreement signals for the selected subset, enabling reproducible analysis of evidence-grounded summarization quality in psychiatry systematic reviews.

\clearpage
\section{Reasoning Trajectory Evaluation}
\label{append:chain_eval_detail}

Given an input context $C$, a reasoning chain
$R=\{r_1,r_2,\dots,r_n\}$, and a final answer $A$, our protocol
evaluates $R$ along three complementary dimensions. These dimensions are
chosen because mental-health AI judgments should be understandable and
auditable, particularly when AI-generated recommendations may influence
clinical decisions \citep{joyce2023explainable,ryan2025assessing}. They capture
observable forms of trajectory-level restraint failure, where the reasoning
does not adequately constrain the specificity or certainty of a judgment to
the available evidence.

\textbf{Usability.} Assesses whether the chain communicates its judgment
clearly and concisely. Terminology Parsimony examines whether clinical terms
are necessary and precise, since excessive language complexity may reduce the
accessibility of mental-health information \citep{aich2025language}. Step
Adequacy examines whether the chain is brief but sufficiently complete:
missing steps may omit evidence, alternatives, or uncertainty, whereas
redundant steps may obscure the decision-relevant content. Both properties are
important for clinician-oriented explanations that can be efficiently reviewed
and appropriately relied upon
\citep{brankovic2025clinician,nicolson2025human}.

\textbf{Logical Structure.} Assesses whether intermediate claims and the
final answer are supported by the input or preceding reasoning units. It
identifies unsupported inferential jumps, irrelevant clinical templates, and
contradictions that indicate an unstable evidence-to-judgment transition. This
distinction is important because fluent or confident medical reasoning does not
necessarily indicate reliable uncertainty assessment or metacognitive
calibration \citep{griot2025large}.

\textbf{Informational Contribution.} Assesses whether each step adds
decision-relevant evidence, qualifications, or alternative interpretations
rather than restating prior content. A useful trajectory should make the basis
and limitations of its judgment inspectable, rather than merely provide a
longer explanation
\citep{brankovic2025clinician,nicolson2025human}.

A chain may satisfy any subset of these dimensions and violate the others. For
example, it may be concise but unsupported, informative but contradictory, or
logically connected yet unnecessarily opaque. We therefore score the three
dimensions independently. These evaluations concern the observable structure
and grounding of the explicit rationale; they do not by themselves establish
clinical correctness, safety, or faithfulness to the model's latent
computation. Scoring details are provided in following sections.

\subsection{Evaluation Settings}
For reproducibility, we will release the complete judge outputs, including dimension-wise scores, relation annotations, and brief rationales. Since the judge prompts and released outputs are model-agnostic, future work can reproduce the protocol with alternative open-source judge models through few-shot prompting or judge-model fine-tuning.

\subsection{LLM-as-Judge Settings and Human Validation}
\label{append:llm_judge_validation}

To support scalable and consistent evaluation of reasoning-trajectory quality, we adopt an LLM-as-judge protocol for all model outputs. Specifically, we use \texttt{gpt-5-mini-2025-08-07} as the evaluation model and provide structured rubrics for Terminology Parsimony, Step Adequacy, Logical Coherence, and Informativeness. The judge is instructed to evaluate each trajectory independently of model identity and to assign ordinal scores according to the criteria described in Section~\ref{append:scoring_metrics}.

To assess the reliability of this automatic evaluation, we conduct a human validation study. The expanded validation set contains 100 reasoning trajectories randomly sampled from the benchmark evaluation outputs. We report the annotation procedure, inter-annotator agreement, LLM-human agreement, and a representative disagreement below.

\subsubsection{Evaluators and Calibration}
\label{append:human_annotators}

The lead evaluator is a certified National Level-II Psychological Counselor who has held the qualification for 14 years and has completed more than 3,000 hours of counseling practice. The evaluator is affiliated with a tertiary Grade-A hospital and serves as a vice president of a municipal psychological association. The lead evaluator developed and calibrated the annotation guidelines and participated in cross-checking and disagreement resolution.

The second evaluator has formal training in psychometrics and received task-specific instruction on the trajectory-evaluation rubrics. Before full annotation, the two evaluators reviewed representative examples and aligned their interpretations of the scoring criteria.

Both evaluators independently annotated the same 100 trajectories using the rubrics for Terminology Parsimony, Step Adequacy, Logical Coherence, and Informativeness. They were instructed to assess whether each trajectory was concise, decision-relevant, structurally supported, and informative, without considering the identity of the model that generated it. Raw agreement was calculated from the independent annotations before discussion. The evaluators then discussed disagreements and produced consensus labels, which serve as the human reference for validating the automatic judge.

\subsubsection{Inter-Annotator Agreement}
\label{append:human_iaa}

Table~\ref{tab:human_iaa} reports raw agreement between the two evaluators on the expanded 100-trajectory validation set. Agreement ranges from 0.83 to 0.92 across the four directly annotated metrics. Terminology Parsimony shows the highest agreement, while Step Adequacy produces more borderline cases because evaluators may differ in whether a reasoning unit is necessary or merely compressible.

\begin{table}[t]
\centering
\small
\setlength{\tabcolsep}{7pt}
\renewcommand{\arraystretch}{1.1}
\caption{Raw inter-annotator agreement on the expanded 100-trajectory human validation set. Agreement is computed from independent annotations before disagreement resolution.}
\label{tab:human_iaa}
\begin{tabular}{lc}
\toprule
Metric & Raw Agreement \\
\midrule
Terminology Parsimony & 0.92 \\
Step Adequacy         & 0.83 \\
Logical Coherence     & 0.85 \\
Informativeness       & 0.86 \\
\bottomrule
\end{tabular}
\end{table}

After independent annotation, disagreements were resolved through discussion to obtain consensus labels. These consensus judgments are used as the human reference for examining the behavior of the automatic evaluator.

\subsubsection{LLM-Human Agreement}
\label{append:llm_human_agreement}

The LLM-human agreement statistics in Table~\ref{tab:llm_human_agreement} are computed on the original pilot subset of 70 trajectories. We report these results separately from the 100-trajectory inter-annotator analysis because the Cohen's $\kappa$ and weighted $\kappa$ values were calculated before the human validation set was expanded. This separation avoids combining agreement statistics obtained from different sample sizes.

For each trajectory in the pilot subset, disagreements between the two human evaluators were resolved through discussion, and the resulting consensus label was compared with the LLM-as-judge score. The automatic evaluator shows perfect agreement with the human consensus for Logical Coherence on this subset and moderate agreement for Terminology Parsimony, Step Adequacy, and Informativeness. These results support the use of automatic scoring for scalable benchmark analysis, while also indicating that borderline usability and informativeness judgments remain sensitive to evaluator interpretation.

\begin{table}[t]
\centering
\small
\setlength{\tabcolsep}{4.5pt}
\renewcommand{\arraystretch}{1.1}
\caption{Agreement between the LLM-as-judge and human consensus labels on the original 70-trajectory pilot subset.}
\label{tab:llm_human_agreement}
\begin{tabular}{lccc}
\toprule
Metric
& Agreement
& Cohen's $\kappa$
& Weighted $\kappa$ \\
\midrule
Terminology Parsimony & 0.743 & 0.426 & 0.489 \\
Step Adequacy         & 0.686 & 0.479 & 0.593 \\
Logical Coherence     & 1.000 & 1.000 & 1.000 \\
Informativeness       & 0.857 & 0.481 & 0.665 \\
\bottomrule
\end{tabular}
\end{table}

\subsubsection{Representative Disagreement}
\label{append:human_disagreement}

A representative disagreement occurred in a cognitive-error classification example in which the speaker described an objectively strong football performance as a ``terrible game.'' Both evaluators agreed that the predicted cognitive-error label was supported by the contrast between the observed performance and the speaker's negative interpretation. However, one evaluator assigned Logical Coherence a score of 2, while the other assigned a score of 1.

The disagreement concerned whether repeating the phrase ``bad game'' merely preserved the speaker's subjective framing or introduced a minor grounding problem by treating that framing as an objective description. After discussion, the evaluators agreed that the core evidence-to-label mapping was correct and that the difference concerned only the strength of one intermediate support relation. This example illustrates that most disagreements arise from borderline judgments about local grounding and step necessity rather than from fundamentally different interpretations of the final conclusion.

\subsection{Scoring Metrics}
\label{append:scoring_metrics}
We adopt a three-level ordinal scale for each metric.

\subsubsection{Usability}
We assess whether the reasoning chain is presented in a clear and efficient manner by measuring \emph{Terminology Parsimony} and \emph{Step Adequacy}, which together reflect unnecessary terminology usage and avoidable verbosity.

\paragraph{Terminology Parsimony.}
\begin{itemize}[leftmargin=1em, itemsep=0pt]
  \item {Low (2 points):}
  Only psychological concepts strictly required for explaining or justifying the decision are used.

  \item {Medium (1 point):}
  Additional background or contextual terminology is introduced but does not hinder comprehension.

  \item {High (0 points):}
  Terminology substantially exceeds decision needs and mainly serves as authority decoration.
\end{itemize}

\paragraph{Step Adequacy.}
\begin{itemize}[leftmargin=1em, itemsep=0pt]
\item {Efficient (2 points):} Every reasoning unit is essential and cannot be removed without weakening the decision rationale.
\item {Acceptable (1 point):} Some content is compressible or could be merged, but the overall reasoning remains understandable and decision-relevant.
\item {Inefficient (0 points):} A substantial portion of sentences can be removed with little or no impact on the decision, indicating verbosity that does not contribute to the conclusion.
\end{itemize}

\subsubsection{Logical Structure} 
We evaluate whether the reasoning forms a coherent and well-connected structure using \emph{Logical Coherence} and a set of \emph{structure-derived submetrics}, which quantify unsupported leaps, irrelevant steps, and contradictions based on relation annotations.

\paragraph{Logical Coherence.}
\begin{itemize}[leftmargin=1em, itemsep=0pt]
\item {Coherent (2 points):}
RCUs are well-formed and most units are meaningfully connected through clear support relations to the context and/or earlier RCUs.

\item {Partially Coherent (1 point):}
RCUs are mostly identifiable, but the chain contains some weakly connected units, minor leaps, or under-specified links.

\item {Incoherent (0 points):}
RCUs are difficult to identify or many units are disconnected, contradictory, or rely on unsupported leaps that undermine the reasoning structure.
\end{itemize}

\paragraph{Structure-derived Submetrics.}
In addition to the three-level coherence score, we compute quantitative submetrics directly from the judge-generated relation annotations in JSON. These submetrics provide fine-grained signals about common structural failure modes and enable scalable benchmark analysis.

Let $\mathcal{U}=\{R_1,\ldots,R_m\}$ denote the set of RCUs. For each $R_i$, the judge outputs a relation label:
\begin{equation}
\resizebox{0.85\linewidth}{!}{%
\ensuremath{
\ell_i \in 
\left\{
\begin{array}{l}
\text{ENTAILS},\ \text{SUPPORTS},\ \text{REFINES},\ \text{CAUSES},\\
\text{CONTRADICTS},\ \text{IRRELEVANT}
\end{array}
\right\}.
}%
}
\end{equation}
We define the following rates:
\begin{equation}
\resizebox{0.9\linewidth}{!}{$
\begin{array}{rcl}
\mathrm{IrrelevanceRate} 
&=& \frac{1}{m}\sum_{i=1}^{m}\mathbb{I}\!\left[\ell_i=\mathrm{IRRELEVANT}\right], \\[4pt]
\mathrm{ContradictionRate} 
&=& \frac{1}{m}\sum_{i=1}^{m}\mathbb{I}\!\left[\ell_i=\mathrm{CONTRADICTS}\right].
\end{array}
$}
\end{equation}
Intuitively, $\mathrm{IrrelevanceRate}$ captures the fraction of steps that do not contribute to the decision, and $\mathrm{ContradictionRate}$ captures explicit conflicts within the reasoning structure. These submetrics complement the coarse coherence score by exposing which failure modes dominate a model's reasoning behavior.

\subsubsection{Informational Contribution}
We assess how effectively the reasoning advances toward the final conclusion using \emph{Step-level Informativeness}, which captures the usefulness of individual steps.

\paragraph{Informativeness.}

\begin{itemize}[leftmargin=1em, itemsep=0pt]
\item {High (2 points):}
Most RCUs are informative and information accumulates toward the answer.

\item {Medium (1 point):}
Some RCUs are informative but redundancy is present.

\item {Low (0 points):}
many RCUs are redundant or irrelevant.
\end{itemize}

\clearpage
\section{\MODEL{} Training Details}
\label{append:mindora_training}

\begin{figure*}[t]
\centering
\includegraphics[width=0.95\linewidth]{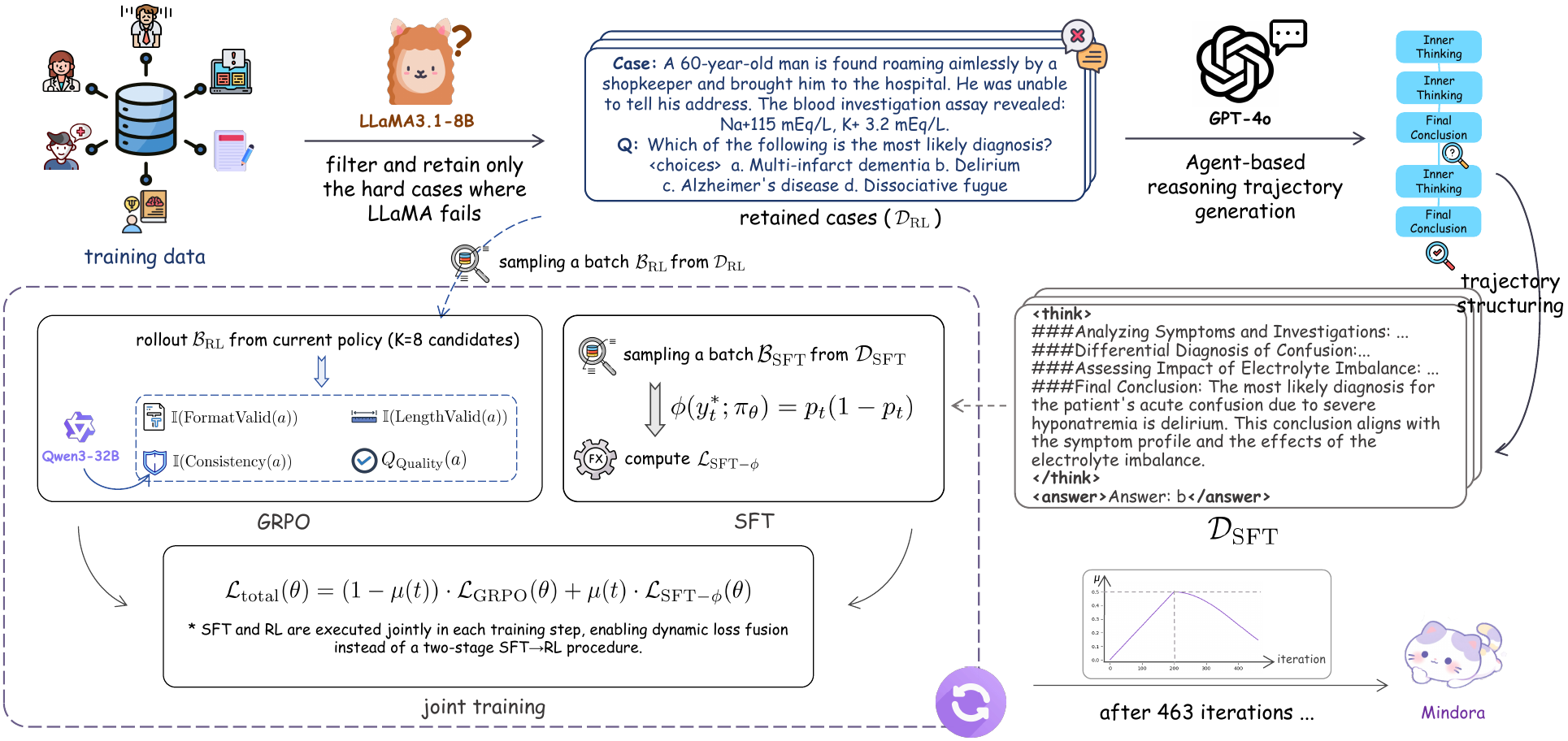}
\caption{\MODEL{} training pipeline: hard-case filtering, verifier-guided reasoning trajectory generation, structured rewriting into non-backtracking trajectories, and hybrid SFT--RL post-training with consistency-aware rewards.}
\label{fig:mindora}
\end{figure*}

\subsection{Overview}

\MODEL{} is a post-trained LLM built on Qwen3-8B and used as a benchmark-guided case study for improving reliable mental-health reasoning. The training pipeline contains two stages. \iffalse The first stage, Reasoning Trajectory Generation (RTG), constructs concise and clinically interpretable chain-of-thought supervision from genuinely difficult cases. \fi {The first stage, Reasoning Trajectory Generation (RTG), constructs concise, interpretable reasoning-trajectory supervision from genuinely difficult cases.} The second stage applies hybrid SFT--RL post-training with a consistency-aware reward, encouraging the model to produce reasoning that is task-relevant, internally consistent, and aligned with the final answer.

\subsection{Reasoning Trajectory Generation}

Effective supervision for mental-health reasoning must be both informative and learnable. Using all available training instances can dilute the signal with easy cases that are solvable through surface cues, while raw search traces can be too verbose or unstable for small-model learning. RTG therefore concentrates supervision on difficult cases and then converts successful trajectories into readable, non-backtracking rationales.

\paragraph{Hard-case filtering.}
We first run zero-shot question answering on each training split using LLaMA-3-8B-Instruct and retain only the instances answered incorrectly. These retained samples are treated as high-signal cases because they are less likely to be solved by superficial pattern matching and more likely to require explicit reasoning. This filtering step aims to improve the density of useful supervision rather than simply increasing training volume.

\paragraph{Verifier-guided search.}
For each retained instance, we generate candidate reasoning trajectories through iterative refinement with a verifier. The search process allows exploration and backtracking to recover a valid solution. We do not use the raw search dynamics as training targets; instead, we retain only the successful trajectory content that supports the correct final answer. This follows standard verifier-guided trajectory construction while keeping the downstream supervision focused on the final usable rationale.

\subsection{Structured Rewriting for Non-backtracking Trajectories}

Raw search traces often contain self-corrections, speculative branches, and backtracking artifacts. While such traces may help a strong teacher model explore the solution space, they can reduce readability and confuse smaller student models during post-training. We therefore rewrite the selected trajectory into a standardized, non-backtracking rationale.

Each rewritten example contains two phases. The reasoning phase is wrapped in \texttt{<think>} and segmented by domain-relevant subtitles, such as \textit{Symptom Analysis}, \textit{Differential Diagnosis}, or \textit{Evidence Assessment}. It ends with a mandatory \textit{Final Conclusion} that explicitly maps the reasoning to the decision. The answer phase is wrapped in \texttt{<answer>} and ends with an unambiguous \texttt{Answer: [option/result]} consistent with the final conclusion.

This rewriting stage enforces three properties. First, it improves linear coherence by presenting a single best path instead of visible exploration. Second, it strengthens reason--answer alignment by requiring the final conclusion and answer field to match. \iffalse Third, it improves interpretability through clinically meaningful modularization. \fi {Third, it improves interpretability through domain-relevant modularization.} This design is particularly important in mental-health settings, where visible self-correction, excessive hedging, or loosely connected clinical language can undermine the clarity of the judgment.

\subsection{Hybrid SFT--RL Post-training}

\MODEL{} is trained with a hybrid SFT--RL scheme that balances supervised imitation and reward-driven exploration. The SFT component encourages the model to imitate the structured reasoning trajectories generated by RTG, while the RL component further refines model behavior according to task-specific correctness and reasoning-quality constraints.

Our emphasis is the consistency-aware reward tailored to mental-health reasoning. The reward combines four components: format validity, length validity, internal consistency, and task-specific correctness. Format validity checks whether the output follows the required \texttt{<think>} and \texttt{<answer>} structure. Length validity discourages degenerate outputs that are either too short to justify the decision or unnecessarily verbose. Internal consistency is evaluated by an auxiliary model, Qwen3-32B, which checks whether the reasoning phase contains factual or logical inconsistencies. Task-specific correctness rewards final answers that match the reference labels or scoring criteria.

The consistency term is designed to penalize reasoning that appears fluent but contains unsupported or self-conflicting intermediate claims. In contrast to RL objectives that primarily optimize final-answer correctness or generic response quality, this reward shapes intermediate rationales toward concise, grounded, and internally consistent trajectories. This makes the training objective aligned with the evaluation target of \BENCH{}, where reliable mental-health reasoning depends not only on the endpoint but also on whether the path to that endpoint remains evidence-proportional and decision-relevant.

\subsection{Role of \MODEL{} in This Work}

\MODEL{} is used as a benchmark-guided post-training case study rather than as a deployable mental-health system. Its purpose is to test whether the failures surfaced by \BENCH{} can motivate targeted optimization toward more reliable reasoning behavior. The model results therefore support the utility of \BENCH{} as both an evaluation tool and a research substrate, while deployment-oriented safety, treatment validity, and clinical efficacy remain outside the scope of this work.

\clearpage
\section{Experimental Details}

\subsection{Evaluated LLMs}
\label{append:baseline_models}
\textbf{} To assess whether model scale, distilled reasoning variants, and the gap between reasoning-oriented and chat-oriented LLMs affect performance on mental health reasoning tasks, we evaluated a broad range of GPT, DeepSeek, Qwen, and LLaMA models. For closed-source systems, we included the reasoning models GPT-o (o4-mini), DeepSeek-R (DeepSeek-R1), and Qwen’s QwQ series (QwQ-plus), as well as the leading chat models GPT-4o, DeepSeek-V3, and Qwen-plus\footnote{Qwen-plus is Qwen’s versatile chat model, also trained with reasoning capabilities.}. For open-source models, we tested Qwen, LLaMA, and DeepSeek-distilled variants across multiple scales, with the full list provided in Table~\ref{tab:main_result}. For psychology-focused LLMs, we evaluated EmoLLM\footnote{\url{https://www.modelscope.cn/models/aJupyter/EmoLLM_Qwen2-7B-Instruct_lora/}.}
, finetuned from Qwen2-7B-Instruct as a well-known mental-health chat model, and Psyche-R1\footnote{\url{https://huggingface.co/MindIntLab/Psyche-R1}}
, an SFT+GRPO-trained Qwen2.5-7B-Instruct model representing the latest mental-health-reasoning LLMs.

\subsection{Model Size and Computational Budget}
\label{append:compute_budget}

For benchmark evaluation, most proprietary, hosted, and large open-weight models are accessed through official APIs, including GPT-o4-mini, GPT-4o, DeepSeek-R1, DeepSeek-V3, Qwen-plus, QwQ-plus, LLaMA-4, LLaMA-3.3-70B, DeepSeek-distilled 70B/32B/14B variants, Qwen2.5-72B, Qwen3-32B/14B, and QwQ-32B. Their exact provider-side compute budgets are not publicly available. The locally evaluated models include EmoLLM, Psyche-R1, and Qwen3-8B, all at the 7--8B scale. Local benchmark inference was conducted on 8 NVIDIA GPUs for approximately 2 hours in total, corresponding to about 16 GPU-hours.

Reasoning-trajectory evaluation is performed with gpt-5-mini-2025-08-07 as an API-based LLM-as-Judge. Combining answer generation through APIs and trajectory evaluation, the total monetary cost for running the benchmark experiments was approximately \$20 under the API pricing used in our experiments. \MODEL{} is initialized from Qwen3-8B and therefore has approximately 8B parameters; its benchmark inference is included in the local compute budget above.

\subsection{Group-wise Performance Analysis}
\label{append:groupwise_analysis}

% rank table
\begin{table*}[htbp]
\centering
\renewcommand{\arraystretch}{1.3}
\caption{Group-wise rank comparison on \BENCH. Lower rank indicates better performance.}
\label{tab:rank_result}
\resizebox{\linewidth}{!}{
\begin{tabular}{l|cccc|cccc|c|ccccc|c|c|c}
\hline
\textbf{Model}
& \makecell[c]{\textbf{Cognitive}\\\textbf{Reframing}}
& \makecell[c]{\textbf{Pattern}\\\textbf{Reframe}}
& \makecell[c]{\textbf{Therapist}\\\textbf{Q\&A}}
& \textcolor{brown}{$Avg_{1}$ Rank}
& \textbf{DepSign}
& \textbf{SWMH}
& \textbf{T-SID}
& \textcolor{brown}{$Avg_{2}$ Rank}
& \textbf{AnnoMI$_{M}$}
& \textbf{MHQA}
& \textbf{MedQA$_{M}$}
& \textbf{MedMCQA$_{M}$}
& \textbf{PubMedQA$_{M}$}
& \textcolor{brown}{$Avg_{4}$ Rank}
& \textbf{PSRS}
& \makecell[c]{\textbf{Mental}\\\textbf{Misinfo}}
& \textcolor{brown}{$Avg_{all}$ Rank} \\
\hline
\multicolumn{18}{c}{\textit{close-source models}} \\ \hline
GPT-o4-mini
& 3 & 10 & 10 & \textcolor{brown}{7.67}
& 27 & 17 & 17 & \textcolor{brown}{20.33}
& 15
& 9 & \cellcolor{lightorange}\textbf{1} & 18 & 19 & \textcolor{brown}{11.75}
& 11
& \cellcolor{lightorange}\textbf{1}
& \textcolor{brown}{12.15} \\
GPT-4o
& 7 & 7 & 7 & \textcolor{brown}{7.00}
& 18 & 12 & 26 & \textcolor{brown}{18.67}
& 7
& 14 & 23 & 21 & 22 & \textcolor{brown}{20.00}
& 7
& 5
& \textcolor{brown}{13.54} \\
Deepseek-R1
& \cellcolor{lightorange}\textbf{1} & \cellcolor{lightorange}\textbf{1} & 8 & \textcolor{brown}{3.33}
& 15 & 8 & 23 & \textcolor{brown}{15.33}
& 18
& 10 & 2 & 2 & 12 & \textcolor{brown}{6.50}
& 9
& 20
& \textcolor{brown}{9.92} \\
Deepseek-V3
& 14 & 5 & 13 & \textcolor{brown}{10.67}
& 9 & 3 & 14 & \textcolor{brown}{8.67}
& 14
& 12 & 4 & 7 & 25 & \textcolor{brown}{12.00}
& 4
& 15
& \textcolor{brown}{10.69} \\
Qwen-plus
& 6 & 14 & 14 & \textcolor{brown}{11.33}
& 19 & 8 & 12 & \textcolor{brown}{13.00}
& 21
& 11 & 3 & 5 & 17 & \textcolor{brown}{9.00}
& 8
& 18
& \textcolor{brown}{12.00} \\
QwQ-plus
& 10 & 2 & 20 & \textcolor{brown}{10.67}
& 15 & 10 & 21 & \textcolor{brown}{15.33}
& 22
& 19 & 21 & 16 & 16 & \textcolor{brown}{18.00}
& 2
& 24
& \textcolor{brown}{15.23} \\
\hline
\multicolumn{18}{c}{\textit{70B+ open-source models}} \\ \hline
LLaMA-4
& 13 & 11 & 16 & \textcolor{brown}{13.33}
& 22 & 22 & 9 & \textcolor{brown}{17.67}
& 16
& 13 & 13 & 4 & 13 & \textcolor{brown}{10.75}
& 23
& 14
& \textcolor{brown}{14.54} \\
LLaMA-3.3-70B
& 17 & 17 & 21 & \textcolor{brown}{18.33}
& 14 & 25 & 24 & \textcolor{brown}{21.00}
& 17
& 20 & 9 & 3 & 11 & \textcolor{brown}{10.75}
& 27
& 5
& \textcolor{brown}{16.15} \\
dsdistill-LLaMA-70B
& 11 & 7 & 18 & \textcolor{brown}{12.00}
& 20 & 11 & 27 & \textcolor{brown}{19.33}
& 11
& 15 & 16 & 20 & 8 & \textcolor{brown}{14.75}
& 10
& 7
& \textcolor{brown}{13.92} \\
Qwen2.5-72B
& 5 & 20 & 17 & \textcolor{brown}{14.00}
& 25 & 7 & 3 & \textcolor{brown}{11.67}
& 8
& 21 & 11 & 14 & 14 & \textcolor{brown}{15.00}
& \cellcolor{lightorange}\textbf{1}
& 16
& \textcolor{brown}{12.46} \\
\hline
\multicolumn{18}{c}{\textit{32B open-source models}} \\ \hline
dsdistill-Qwen32B
& 14 & 6 & 22 & \textcolor{brown}{14.00}
& 21 & 18 & 20 & \textcolor{brown}{19.67}
& 12
& 22 & 20 & 24 & 14 & \textcolor{brown}{20.00}
& 6
& 13
& \textcolor{brown}{16.31} \\
Qwen3-32B
& 18 & 14 & 12 & \textcolor{brown}{14.67}
& 25 & 5 & 5 & \textcolor{brown}{11.67}
& 19
& 16 & 12 & 17 & 21 & \textcolor{brown}{16.50}
& 3
& 23
& \textcolor{brown}{14.62} \\
QwQ-32B
& 7 & 3 & 18 & \textcolor{brown}{9.33}
& 15 & 12 & 22 & \textcolor{brown}{16.33}
& 23
& 17 & 19 & 12 & 10 & \textcolor{brown}{14.50}
& 4
& 26
& \textcolor{brown}{14.46} \\
\hline
\multicolumn{18}{c}{\textit{14B open-source models}} \\ \hline
dsdistill-Qwen14B
& 12 & 23 & 23 & \textcolor{brown}{19.33}
& 11 & 16 & 11 & \textcolor{brown}{12.67}
& 9
& 24 & 22 & 25 & 7 & \textcolor{brown}{19.50}
& 12
& 19
& \textcolor{brown}{16.46} \\
Qwen3-14B
& 14 & 11 & 5 & \textcolor{brown}{10.00}
& 23 & 14 & 16 & \textcolor{brown}{17.67}
& 24
& 18 & 17 & 18 & 24 & \textcolor{brown}{19.25}
& 16
& 10
& \textcolor{brown}{16.15} \\
\hline
\multicolumn{18}{c}{\textit{7$\sim$8B open-source models}} \\ \hline
LLaMA3.1-8B
& 25 & 24 & 27 & \textcolor{brown}{25.33}
& 6 & 18 & 14 & \textcolor{brown}{12.67}
& 13
& 27 & 25 & 23 & 20 & \textcolor{brown}{23.75}
& 26
& 9
& \textcolor{brown}{19.77} \\
dsdistill-LLaMA-8B
& 24 & 25 & 25 & \textcolor{brown}{24.67}
& 10 & 23 & 19 & \textcolor{brown}{17.33}
& 20
& 26 & 26 & 26 & 6 & \textcolor{brown}{21.00}
& 15
& 2
& \textcolor{brown}{19.00} \\
EmoLLM
& 19 & 26 & 24 & \textcolor{brown}{23.00}
& 7 & 24 & 8 & \textcolor{brown}{13.00}
& 27
& 25 & 27 & 27 & 26 & \textcolor{brown}{26.25}
& 21
& 11
& \textcolor{brown}{20.92} \\
Psyche-R1
& 27 & 27 & 26 & \textcolor{brown}{26.67}
& 11 & 27 & 7 & \textcolor{brown}{15.00}
& 10
& 5 & 18 & 15 & 18 & \textcolor{brown}{14.00}
& 14
& 4
& \textcolor{brown}{16.08} \\
Qwen3-8B
& 23 & 14 & 14 & \textcolor{brown}{17.00}
& 11 & 18 & 17 & \textcolor{brown}{15.33}
& 25
& 23 & 24 & 22 & 23 & \textcolor{brown}{23.00}
& 17
& 25
& \textcolor{brown}{19.69} \\
\hline
Mindora$_{SFT}$
& 26 & 20 & 8 & \textcolor{brown}{18.00}
& 2 & 15 & 13 & \textcolor{brown}{10.00}
& 4
& 7 & 15 & 10 & 9 & \textcolor{brown}{10.25}
& 22
& 22
& \textcolor{brown}{13.31} \\
Mindora$_{RL}$
& 20 & 17 & 11 & \textcolor{brown}{16.00}
& 24 & 18 & 25 & \textcolor{brown}{22.33}
& 26
& 6 & 5 & \cellcolor{lightorange}\textbf{1} & 5 & \textcolor{brown}{4.25}
& 18
& 27
& \textcolor{brown}{15.62} \\
Mindora$_{SFT+RL}$
& 22 & 20 & 6 & \textcolor{brown}{16.00}
& \cellcolor{lightorange}\textbf{1} & 3 & 2 & \textcolor{brown}{2.00}
& 6
& 3 & 6 & 9 & 3 & \textcolor{brown}{5.25}
& 24
& 21
& \textcolor{brown}{9.69} \\
Mindora w/o S1
& 4 & 13 & \cellcolor{lightorange}\textbf{1} & \textcolor{brown}{6.00}
& 8 & 26 & 4 & \textcolor{brown}{12.67}
& \cellcolor{lightorange}\textbf{1}
& 4 & 8 & 6 & 4 & \textcolor{brown}{5.50}
& 19
& 12
& \textcolor{brown}{8.46} \\
Mindora w/o S2
& 21 & 19 & 2 & \textcolor{brown}{14.00}
& 5 & 2 & \cellcolor{lightorange}\textbf{1} & \textcolor{brown}{2.67}
& 2
& 8 & 14 & 13 & 27 & \textcolor{brown}{15.50}
& 20
& 17
& \textcolor{brown}{11.62} \\
Mindora w/o S3
& 9 & 9 & 2 & \textcolor{brown}{6.67}
& 3 & 5 & 10 & \textcolor{brown}{6.00}
& 3
& 2 & 7 & 8 & \cellcolor{lightorange}\textbf{1} & \textcolor{brown}{4.50}
& 25
& 8
& \textcolor{brown}{7.08} \\
Mindora
& 2 & 3 & 4 & \cellcolor{lightorange}\textcolor{brown}{\textbf{3.00}}
& 4 & \cellcolor{lightorange}\textbf{1} & 6 & \textcolor{brown}{3.67}
& 5
& \cellcolor{lightorange}\textbf{1} & 10 & 10 & 2 & \textcolor{brown}{5.75}
& 13
& 3
& \cellcolor{lightorange}\textcolor{brown}{\textbf{4.92}} \\
\hline
\end{tabular}
}
\end{table*}

To complement the absolute performance scores in Table~\ref{tab:main_result}, we further conduct a group-wise rank analysis in Table~\ref{tab:rank_result}. This analysis is intended to reduce the influence of any single high-scoring dataset and to examine whether a model performs consistently across heterogeneous mental-health reasoning tasks. For each dataset, models are ranked according to task performance, with lower ranks indicating better performance. We then compute average ranks within each task group, including appraisal-oriented cognitive error identification ($Avg_1$), mental-condition detection ($Avg_2$), multi-step psychiatry QA ($Avg_4$), as well as the overall average rank across all evaluated datasets.

The results show that Mindora achieves the best overall average rank ($Avg_{all}$ Rank = 4.92), outperforming both strong proprietary models and domain-specialized baselines. This indicates that Mindora's advantage is not driven by a single dataset or task family, but reflects more stable performance across diverse reasoning settings. In comparison, DeepSeek-R1 obtains strong ranks on appraisal tasks ($Avg_1$ Rank = 3.33) and multi-step psychiatry QA ($Avg_4$ Rank = 6.50), but performs less consistently on mental-condition detection and misinformation identification, leading to a weaker overall rank ($Avg_{all}$ Rank = 9.92). GPT-o4-mini ranks first on MedQA$_M$ and MentalMisinfo, but its relatively weak ranks on diagnosis-oriented datasets result in a higher overall average rank ($Avg_{all}$ Rank = 12.15).

The group-wise view also reveals that different model families exhibit distinct strengths. Large proprietary and reasoning-oriented models remain competitive on structured QA and verification tasks, whereas their ranks fluctuate more substantially on narrative-based appraisal and diagnosis tasks. By contrast, Mindora maintains consistently strong ranks across appraisal ($Avg_1$ Rank = 3.00), diagnosis ($Avg_2$ Rank = 3.67), and multi-step reasoning ($Avg_4$ Rank = 5.75), while also ranking near the top on MentalMisinfo. These results support the conclusion that targeted post-training improves not only pooled task accuracy, but also cross-task robustness over heterogeneous mental-health reasoning scenarios.

The ablation variants further clarify the contribution of individual components. Mindora w/o S3 remains competitive overall ($Avg_{all}$ Rank = 7.08), but its weaker ranks on PSRS and MentalMisinfo suggest that the consistency reward is particularly important for tasks requiring stable evidence integration and verification. Mindora w/o S2 performs well on mental-condition detection ($Avg_2$ Rank = 2.67), but its weaker multi-step rank ($Avg_4$ Rank = 15.50) indicates that structured rewriting is important for transferring reasoning patterns \iffalse across more complex clinical QA settings \fi {across more complex mental-health QA tasks}. Overall, the rank analysis is consistent with the absolute-score results and reinforces that Mindora provides the most balanced performance across the benchmark.

\subsection{External Evaluation on GoEmotions}
\label{append:external_evaluation}

To examine whether the improvements of \MODEL{} extend beyond the datasets and task formats included in \BENCH{}, we conduct an external evaluation on GoEmotions~\citep{demszky2020goemotions}. GoEmotions is a manually annotated dataset for fine-grained emotion classification. We evaluate \MODEL{} and its Qwen3-8B backbone on 1,012 test instances using classification accuracy. \MODEL{} achieves an accuracy of 0.4407, compared with 0.3528 for Qwen3-8B. Although GoEmotions does not evaluate the full scope of mental-health reasoning, this result provides complementary evidence that the gains of \MODEL{} are not confined to the specific datasets used for its post-training and evaluation.

\paragraph{Case Study.}
Figure~\ref{fig:case_study_goemo} illustrates a case in which both models predict the correct emotion labels, \textit{annoyance} and \textit{disapproval}.

Both models reach the correct answer, but their trajectories differ in selectivity. \MODEL{} directly connects the absence of constructive counter-arguments to the expressed frustration and criticism. Qwen3-8B repeats the same evidence, eliminates several unsupported alternatives, and restates the label comparison. This example illustrates that identical final-answer correctness can conceal differences in trajectory length, evidence selection, and informational contribution.

\clearpage
\clearpage \begin{figure*}[p]
\centering
\begin{minipage}{0.98\textwidth}
\begin{tcolorbox}[width=\textwidth]
\textbf{GoEmotions Case Content:}\\[2pt]
{\small
It's kinda odd that people are downvoting you without providing well-sourced counter-arguments to the points you're making.
}
\end{tcolorbox}

\begin{tcolorbox}[width=\textwidth]
\textbf{Response of Mindora:}\\[2pt]
{\small
\#\#\#{Understanding the Context:} The comment discusses downvoting without counter-arguments, suggesting frustration with the lack of constructive feedback.\\
\#\#\#{Analyzing Emotional Tone:} The tone implies dissatisfaction with the downvoting process, indicating a negative sentiment.\\
\#\#\#{Identifying Emotional Expression:} The frustration and dissatisfaction suggest emotions such as annoyance and disapproval.\\
\#\#\#{Final Conclusion:} The emotions expressed in the comment are annoyance and disapproval.
}
\end{tcolorbox}

\begin{tcolorbox}[width=\textwidth]
\textbf{Response of Qwen3-8B:}\\[2pt]
{\small
\#\#\#{Understanding the Context:} The comment expresses frustration with people downvoting without providing well-reasoned counter-arguments. This indicates a negative reaction to the lack of constructive engagement.\\
\#\#\#{Identifying Emotions:}\\
- {Annoyance:} The phrase ``kinda odd'' suggests mild irritation or frustration with the downvoting behavior.
\\
- {Disapproval:} The speaker is critical of downvoting without counter-arguments, indicating a negative judgment of the behavior.
\\
{Eliminating Other Emotions:} There is no clear indication of admiration, approval, joy, or other positive emotions. The tone is neutral to negative, with no signs of confusion, curiosity, or surprise.
\\
\#\#\#{Comparing with the Given Choices:} Annoyance is supported by the speaker's frustration with the downvoting behavior, while disapproval is reflected in the criticism of the lack of counter-arguments. Other emotions such as confusion or curiosity are not clearly indicated.
\\
\#\#\#{Final Conclusion:} The emotions expressed in the comment are annoyance and disapproval.
}
\end{tcolorbox}

\caption{Reasoning trajectory comparison between Mindora and Qwen3-8B on GoEmotions.}
\label{fig:case_study_goemo}
\end{minipage}
\end{figure*}

\clearpage
\begin{table*}[ht]
\centering
\renewcommand{\arraystretch}{1.3}
\caption{Reasoning-trajectory evaluation conditioned on correct final predictions. Unlike Table~\ref{tab:trajectory_main}, incorrect predictions are excluded rather than assigned zero trajectory scores. Scores are therefore not directly comparable in magnitude with the unconditional averages in Table~\ref{tab:trajectory_main}.}
\label{tab:trajectory_correct_only}
\resizebox{0.8\linewidth}{!}{
\begin{tabular}{lccccccc}
\hline
\textbf{Model}
& \makecell[c]{\textbf{Terminology}\\\textbf{Parsimony $\uparrow$}}
& \makecell[c]{\textbf{Step}\\\textbf{Adequacy $\uparrow$}}
& \makecell[c]{\textbf{Avg Steps$\downarrow$}}
& \makecell[c]{\textbf{Logical}\\\textbf{Coherence $\uparrow$}}
& \makecell[c]{\textbf{Irrelevance}\\\textbf{Rate $\downarrow$}}
& \makecell[c]{\textbf{Contradiction}\\\textbf{Rate $\downarrow$}}
& \textbf{Informativeness} $\uparrow$ \\
\hline
GPT-o4-mini
& 1.8571
& \textbf{1.9238}
& \textbf{4.8762}
& 1.8762
& 0.0145
& 0.0231
& 1.5810 \\

DeepSeek-R1
& 1.4202
& 1.3109
& 7.4118
& \textbf{1.9916}
& 0.0249
& 0.0828
& 1.8824 \\

Qwen-plus
& 1.3178
& 0.9720
& 9.0093
& 1.8505
& 0.0643
& 0.0590
& \textbf{1.9159} \\

LLaMA-4
& \textbf{1.8761}
& 1.8673
& 4.9558
& 1.8319
& 0.0196
& \textbf{0.0105}
& 1.5664 \\

Psyche-R1
& 1.6275
& 1.4314
& 5.5882
& 1.9216
& 0.0173
& 0.0538
& 1.6765 \\

Qwen3-8B
& 1.4112
& 0.8785
& 7.6168
& 1.7850
& 0.0556
& 0.0480
& 1.9065 \\
\hline

\MODEL{}
& 1.5873
& 1.5238
& 5.0079
& 1.8810
& \textbf{0.0124}
& 0.0293
& 1.7619 \\

\MODEL{} w/o S1
& 1.6396
& 1.3333
& 5.9910
& 1.8829
& 0.0129
& 0.0352
& 1.7207 \\

\MODEL{} w/o S2
& 0.9915
& 0.3644
& 9.8136
& 1.8475
& 0.0326
& 0.0778
& 1.8898 \\

\MODEL{} w/o S3
& 1.6102
& 1.5339
& 4.9576
& 1.8559
& 0.0158
& 0.0233
& 1.7203 \\
\hline
\end{tabular}%
}
\end{table*}

\subsection{Trajectory Evaluation Conditioned on Correct Predictions}
\label{append:correct_conditioned_trajectory}

The primary trajectory evaluation assigns zero scores to trajectories associated with incorrect final answers. Although this design jointly reflects endpoint correctness and trajectory quality, it may confound differences in reasoning quality with differences in task accuracy. We therefore conduct an additional evaluation that includes only instances for which each model produces the correct final prediction.

Table~\ref{tab:trajectory_correct_only} reports the conditioned results. \MODEL{} produces substantially shorter trajectories than its Qwen3-8B backbone, with an average of 5.0079 steps compared with 7.6168. It also improves Step Adequacy from 0.8785 to 1.5238 and reduces the Irrelevance Rate from 0.0556 to 0.0124. The two models achieve comparable Informativeness, indicating that the reduction in length and irrelevant content does not result from removing decision-relevant information.

The ablation results further demonstrate the contribution of structured trajectory rewriting. Without this component, the average trajectory length increases from 5.0079 to 9.8136 steps, Step Adequacy decreases from 1.5238 to 0.3644, and the Irrelevance and Contradiction Rates increase from 0.0124 and 0.0293 to 0.0326 and 0.0778, respectively. Thus, the trajectory improvements remain observable after controlling for final-answer correctness and cannot be attributed solely to the higher task accuracy of \MODEL{}.

\clearpage
\section{Case Study}
\label{append:general_case_study}

Figures~\ref{fig:case_study_all1}--\ref{fig:case_study_all3} present a representative CognitiveReframing instance where the correct label (\textit{Overgeneralization}) hinges on a minimal discriminative mapping: a \emph{single negative incident} is extrapolated into a \emph{global belief} about others' intentions. While most models can recover the label, their reasoning trajectories exhibit different trade-offs between readability, structure, and decision relevance. 

{Mindora} provides the most balanced trace: it remains anchored to the client’s stated \emph{thought}, isolates the core cue with minimal redundancy, and performs an explicit evidence-to-label mapping under a stable multi-step template, yielding a trajectory that is simultaneously interpretable to users and well-posed as supervision for smaller models. In contrast, several baselines tend to over-index on a single axis: some favor brevity via generic ``key analysis'' templates that can become under-specified in harder cases, while others adopt exhaustive elimination or option-scanning that increases length and cognitive load without improving the discriminative justification. More generally, these trajectories either introduce compressible content (verbosity without additional decision signal) or omit critical linking steps (insufficient mapping from evidence to category), whereas Mindora better preserves a consistent structure that supports both human readability and reliable distillation.

\begin{figure*}[t]
\centering
\begin{minipage}{0.98\textwidth}
\begin{tcolorbox}[width=\textwidth]
\textbf{CognitiveReframing Case Content:}\\[2pt]
{\small
Question: Please choose the most proper cognitive error of this client from the following list: ...\\
Context: \\
- situation: Someone pushed me and made me bleed.\\
- thought: I feel like people out there try to take advantage of me and also betray me.\\
}
\end{tcolorbox}

\begin{tcolorbox}[width=\textwidth]
\textbf{Response of Mindora:}\\[2pt]
{\small
\#\#\#Analyzing Thought Pattern:\\
The client's thought pattern suggests a belief that people generally try to take advantage of and betray them, based on the experience of being pushed and bleeding.\\
\#\#\#Evaluating Cognitive Errors:\\
The thought pattern reflects a belief that people are inherently trying to harm or betray them. This suggests a cognitive error that involves making broad assumptions about others based on specific experiences.\\
\#\#\#Matching Cognitive Error:\\
The cognitive error that involves making broad assumptions about others based on specific experiences is 'Overgeneralization'. The client is generalizing the experience of being pushed to a belief that people generally try to take advantage of them.\\
\#\#\#Final Conclusion:\\
The most proper cognitive error for this client is 'Overgeneralization', as they are generalizing their experience to a broader belief about others.
}
\end{tcolorbox}

\caption{Reasoning trajectory comparison between Mindora and baseline models (1).}
\label{fig:case_study_all1} \end{minipage} \end{figure*}

% cognitive_reframing_92
\begin{figure*}[t]
\centering
\begin{tcolorbox}[width=0.98\textwidth]
\textbf{Response of GPT-o4-mini:}\\[2pt]
{\small
1. Key Analysis:\\
The client experienced a single aggressive incident (being pushed and bleeding) and now believes broadly that all people intend to take advantage of or betray them.\\
2. Step Reasoning:\\
• One negative event → broad conclusion about everyone\\
• Generalizing from a single experience  \\
3. Final Answer:\\
Answer: Overgeneralization
}
\end{tcolorbox}

\begin{tcolorbox}[width=0.98\textwidth]
\textbf{Response of Deepseek-R1:}\\[2pt]
{\small
Based on the provided situation and thought, the client's cognitive error is Overgeneralization.\\
- Situation: A specific event where one person pushed them, causing injury.\\
- Thought: I feel like people out there try to take advantage of me and also betray me.\\
- Error: The client is taking the negative action of one specific person in one specific situation (being pushed) and generalizing it to all people (people out there). They conclude that this is a universal pattern (people trying to take advantage and betray them) based on this single incident. This fits the definition of Overgeneralization: generalize a particular event leading to a certain outcome to all similar events (or in this case, to all people).\\
Why other options are less fitting:\\
- Catastrophizing: Focuses on worst outcomes, not generalizing people's intent.\\
- Discounting the Positive: No positive event mentioned to discount.\\
- Personalization: Client isn't blaming themselves; they blame others.\\
- All-or-Nothing Thinking: Thought isn't about extremes of good/bad regarding a specific thing or person.\\
- Mental Filtering: Only one event is described; no other aspects are mentioned to filter out.\\
- Mind Reading: Client isn't inferring the specific pusher's thoughts; they are generalizing about all people's intentions.\\
- Fortune Telling: Client isn't predicting a future outcome; they are making a present generalization about people's character/intent.\\
- Should Statements: No demands for behavior are present in the thought.\\
- Labeling and Mislabeling: While related, labeling typically attributes the action to the character of the specific person involved (e.g., That person who pushed me is a monster). Here, the client is labeling all people (people out there) based on the action of one, which is the core of overgeneralization.\\
Answer: [Overgeneralization]
}
\end{tcolorbox}

\begin{tcolorbox}[width=0.98\textwidth]
\textbf{Response of Qwen-plus:}\\[2pt]
{\small
1. Key Analysis: The client experienced a single negative event (being pushed) and generalized it to believe all people are harmful and untrustworthy.  \\
2. Step Reasoning:  \\
  - The client extrapolates one negative experience to an overarching belief about others.  \\
  - This aligns with drawing broad conclusions from isolated incidents (overgeneralization).  \\
3. Final Answer: Answer: [Overgeneralization]
}
\end{tcolorbox}

\caption{Reasoning trajectory comparison between Mindora and baseline models (2).}
\label{fig:case_study_all2}
\end{figure*}

% cognitive_reframing_92
\begin{figure*}[!t]
\centering

\begin{tcolorbox}[width=0.98\textwidth]
\textbf{Response of LLaMA-4:}\\[2pt]
{\small
Key Analysis: The client feels that people are trying to take advantage of and betray them after being pushed and hurt. This suggests a pattern of negative thinking and attributing malicious intent to others based on a single incident.\\
Step Reasoning:\\
- The client's experience is a single incident of being pushed and hurt.\\
- The client generalizes that people try to take advantage of and betray them.\\
- This generalization seems to stem from jumping to conclusions based on one experience.\\
Final Answer: Answer: [Overgeneralization]
}
\end{tcolorbox}

\begin{tcolorbox}[width=0.98\textwidth]
\textbf{Response of Psyche-R1:}\\[2pt]
{\small
</think>\\
\#\#\#Comparing with Given Choices:\\
- All-or-Nothing Thinking: The client does not seem to be thinking in extreme categories.\\
- Overgeneralization: The client is drawing a broad conclusion from a single event.\\
- Labeling: The client does not appear to be defining himself based on one action or characteristic.\\
- Fortune Telling: The client does not seem to be predicting the future.\\
- Mind Reading: The client does not assume he knows what others are thinking.\\
- Emotional Reasoning: The client’s thoughts are not directly treated as facts.\\
- Should Statements: The client does not set unrealistic expectations for himself.\\
- Personalization: The client does not take the event personally.\\
- Discounting the Positive: The client does not ignore positive events.\\
- Catastrophizing: The client does not focus on the worst--- Case scenario.\\
- Comparing and Despairing: The client does not compare his worst to someone else's best.\\
- Blaming: The client does not give away his own power to others.\\
- Negative Feeling or Emotion: The client is stuck on a distressing thought, but this is more a symptom than a cognitive error.\\
\#\#\#Final Conclusion: The most proper cognitive error of this client is Overgeneralization.\\
</think>\\
Answer: [Overgeneralization]
}
\end{tcolorbox}

\begin{tcolorbox}[width=0.98\textwidth]
\textbf{Response of Qwen3-8B:}\\[2pt]
{\small
Key Analysis: The client generalizes a single negative event (being pushed) to a broader belief about others' intentions. \\ 
Step Reasoning:  \\
- The incident (being pushed) is a single experience.  \\
- The client concludes people "try to take advantage" and "betray," implying a universal pattern.  \\
- This matches jumping to a broad conclusion from one event.\\ 
Final Answer: [Overgeneralization]
}
\end{tcolorbox}

\caption{Reasoning trajectory comparison between Mindora and baseline models (3).}
\label{fig:case_study_all3}
\end{figure*}

\clearpage
\section{Ablation Details}
\label{append:ablation_details}
We conduct three ablations to isolate the contributions of S1 (hard-sample filtering), S2 (structured rewriting), and S3 (consistency reward). Removing any component consistently degrades average \textsc{MiraMind} performance, confirming that Mindora’s gains do not come from a single training trick but from complementary design choices.

S1 primarily improves robustness on challenging, decision-sensitive instances: replacing hard-case filtering with random sampling reduces overall scores and yields less efficient trajectories, with lower step adequacy and informativeness despite often reaching similar final answers. S2 contributes the largest drop when removed, indicating that rewriting into a standardized, modular format is critical for transferring \iffalse clinically aligned reasoning \fi {benchmark-aligned reasoning} into an 8B-scale student; without S2, trajectory quality sharply deteriorates and inconsistency (irrelevance/contradiction) increases.
S3 further stabilizes reasoning reliability, with the clearest impact on verification/abstraction tasks that require consistent evidence integration (e.g., PSRS and MentalMisinfo). Without S3, logical coherence and informativeness decline and irrelevant steps increase, suggesting the reward is effective at discouraging internally inconsistent or weakly grounded intermediate rationales. Detailed analyses and case studies are provided below.

\subsection{Hard-sample Filtering (S1)}
We examine the effect of hard-case filtering (S1) by comparing the full Mindora model with Mindora w/o S1, where training samples are randomly selected instead of being filtered by zero-shot failures. Removing S1 leads to a consistent degradation in overall performance, with the average score across \textsc{MiraMind} dropping (−0.0202). The performance gap is more pronounced on diagnosis tasks (−0.0358) and \iffalse multi-step clinical reasoning tasks \fi {multi-step psychiatry QA tasks} (−0.0093), indicating weakened generalization to challenging and decision-sensitive cases.

Reasoning-chain evaluation further reveals a systematic quality gap. Compared with Mindora, Mindora w/o S1 shows substantially lower step adequacy (−0.1667) and informativeness (−0.1042), alongside reduced terminology economy (−0.0584). Although both models often reach the same final answer, the variant without S1 tends to produce longer and less efficient reasoning chains, with additional steps that contribute limited new information.

Overall, these results indicate that S1 improves not only task accuracy but also the efficiency and structure of reasoning trajectories. By concentrating training on genuinely difficult cases, hard-case filtering encourages Mindora to learn concise, discriminative, and decision-oriented reasoning patterns rather than verbose or weakly informative explanations. A detailed case study is provided below.

\noindent\textbf{Case Study of Hard-case Filtering (S1).}
\label{append:case_ablation_s1}
To further illustrate the effect of hard-case filtering, we present a representative example from the mental health condition detection task, where both models predict the correct label (none), yet differ markedly in reasoning style. The full responses are shown in Figure~\ref{fig:s1_ablation}.

\begin{figure*}[t]
\centering

\begin{tcolorbox}[width=0.98\textwidth]
\textbf{T-SID Case Content:}\\[2pt]
{\small
Question: Please choose the most proper mental disorder of this post from: suicide, depression, ptsd, or none. Where 'none' means the poster shows no sign of mental disorder. \\
Context: @m\_jxzFVNeQM I would like this specialist centre up (Chy Sawel) \&amp; running and there is interest from professionals willing to get involved.}
\end{tcolorbox}

\begin{tcolorbox}[width=0.98\textwidth]
\textbf{Response of Mindora:}\\[2pt]
{\small
<think>\\
\#\#\#Understanding the Context:\\
The context involves a desire to establish a specialist center and interest from professionals. This suggests a focus on a specific goal or project rather than personal emotional issues.\\
\#\#\#Evaluating Emotional Indicators:\\
The statement lacks emotional indicators such as sadness, hopelessness, or trauma, which are typical signs of depression, suicide, or PTSD.\\
\#\#\#Assessing for Signs of Mental Disorders:\\
There are no explicit mentions of symptoms related to depression, suicide, or PTSD. The focus is on professional involvement and establishing a center.\\
\#\#\#Final Conclusion:\\
Based on the context and lack of emotional indicators, the most proper classification for the poster is `none', indicating no sign of mental disorder.\\
</think>\\
<answer>Answer: none</answer>
}
\end{tcolorbox}

\begin{tcolorbox}[width=0.98\textwidth]
\textbf{Response of Mindora w/o S1 (i.e. the random sampling version):}\\[2pt]
{\small
<think>\\
\#\#\#Understanding Context:\\
The context provided is a tweet expressing a desire for a specialist center to be established and mentions interest from professionals willing to get involved. This suggests a focus on professional and community involvement rather than personal mental health issues.\\
\#\#\#Analyzing for Signs of Mental Disorders:\\
To determine if the poster shows signs of a mental disorder, we need to look for indicators of suicide, depression, or PTSD. The tweet does not mention any personal struggles, emotional distress, or traumatic experiences that would suggest these disorders.\\
\#\#\#Evaluating for `None':\\
The tweet is focused on professional and community initiatives, indicating no personal mental health concerns. Therefore, `none' might be appropriate if the poster shows no signs of mental disorder.\\
\#\#\#Considering Potential Misinterpretations:\\
While the tweet is about a professional initiative, it is important to ensure that there is no underlying personal mental health issue that is indirectly indicated. The tweet does not provide any such clues.\\
\#\#\#Final Conclusion:\\
The poster does not show any signs of mental disorders such as suicide, depression, or PTSD. The tweet is focused on professional and community involvement, suggesting `none' is the most appropriate classification.\\
</think>\\
<answer>Answer: none</answer>
}
\end{tcolorbox}

\caption{Reasoning trajectory comparison between Mindora and Mindora w/o S1.}
\label{fig:s1_ablation}
\end{figure*}

The final Mindora model produces a compact and well-structured reasoning chain. Each step introduces a distinct decision-relevant check: (1) clarifying the task context, (2) examining the presence of emotional indicators, and (3) explicitly ruling out diagnostic criteria for depression, suicide, and PTSD. The reasoning terminates once sufficient evidence has been accumulated to justify the none classification. Notably, no speculative detours or meta-level uncertainty checks are introduced. This trajectory exemplifies high step adequacy and information efficiency, where every reasoning unit contributes directly to narrowing the decision space.

% In contrast, the model trained without hard-case filtering exhibits a noticeably more verbose pattern. Although the content is factually correct, the reasoning chain repeatedly restates the absence of symptoms and introduces an additional meta-level step (“Considering Potential Misinterpretations”) that does not add new evidence. This results in a longer chain with partially redundant reasoning units. From a trajectory-quality perspective, these extra steps increase cognitive load without improving decision clarity, reflecting lower informativeness and weaker compressibility.

% This case study demonstrates that hard-case filtering does not primarily change what the model predicts, but how it reasons. By focusing training on difficult samples, Mindora learns to stop reasoning once the decision is sufficiently supported, avoiding unnecessary elaboration and speculative padding. Such behavior is particularly important in mental-health settings, where over-explanation or excessive hedging can obscure the core judgment and reduce interpretability for end users.
In contrast, the model trained without hard-case filtering is factually correct but noticeably more verbose. Its trajectory repeatedly restates symptom absence and adds a meta-level step (“Considering Potential Misinterpretations”) without introducing new evidence, resulting in redundant reasoning units and lower compressibility. This case suggests that hard-case filtering mainly changes how the model reasons rather than what it predicts: by focusing on difficult samples, Mindora learns to stop once the decision is sufficiently supported, avoiding unnecessary elaboration and speculative padding. This is important in mental-health settings, where over-explanation can obscure the core judgment and reduce interpretability.

\clearpage
\subsection{Structured Rewriting (S2)}
We next analyze S2 (structured rewriting), which is primarily designed to improve the readability and pedagogical quality of training trajectories. Compared with the full model, removing S2 yields a substantial drop in overall performance (−0.0468). The degradation is even more pronounced in trajectory-level evaluation: Step Adequacy sharply declines (−0.6708), and Terminology Economy decreases (−0.3500). In addition, structural reliability deteriorates, as reflected by higher Irrelevance Rate (+0.0095) and Contradiction Rate (+0.0229). These results indicate that training with non-rewritten, unstructured reasoning traces tends to introduce compressible or weakly connected content, increasing cognitive burden without strengthening decision support.

Importantly, S2 also affects the teacher quality of supervision. While strong reasoning models may still succeed with unstructured trajectories, such traces can be suboptimal for distillation into smaller models: they often contain backtracking artifacts and loosely organized explanations, which provide weaker step boundaries and less stable alignment between intermediate rationale and the final decision. By enforcing a standardized, modular format with an explicit final conclusion, S2 converts exploratory traces into consistent and learnable supervision signals, which are critical for transferring \iffalse clinically aligned reasoning \fi {benchmark-aligned reasoning} patterns to an 8B-scale student model. Notably, although Informativeness is slightly higher without S2 (+0.0125), this gain coincides with large drops in adequacy and increased inconsistency, suggesting that additional content is not necessarily more useful when it is not organized into decision-relevant reasoning steps. A detailed case study is provided below.

\noindent \textbf{Case Study of Structured Rewriting (S2).}
\label{append:case_ablation_s2}
To illustrate why S2 is critical for trajectory readability and distillation quality, we compare the full Mindora output with the unstructured variant on a representative cognitive error identification instance, as is shown in Figure~\ref{fig:s2_ablation}. Both models predict the same label (personalization), but the reasoning traces exhibit substantially different instructional properties.

\begin{figure*}[t]
\centering

\begin{tcolorbox}[width=0.98\textwidth]
\textbf{CognitiveReframing Case Content:}\\[2pt]
{\small
Question: Please choose the most proper cognitive error of this client from the following list: ...\\
Context: \\
- situation: I got a parking ticket because my sister forgot to register for overnight parking in front of my house.\\
- thought: I am responsible for her mistakes.}
\end{tcolorbox}

\begin{tcolorbox}[width=0.98\textwidth]
\textbf{Response of Mindora:}\\[2pt]
{\small
\#\#\#Understanding the Thought:\\
The thought expressed is \textit{I am responsible for her mistakes.} This suggests that the individual feels a sense of accountability for the actions of their sister.\\
\#\#\#Identifying Cognitive Error: \\
The thought process involves taking responsibility for the sister's mistake, which aligns with the cognitive error of \textit{Personalization}. Personalization is when one takes things personally or makes them about oneself, even when the responsibility does not lie with them.\\
\#\#\#Final Conclusion: \\
The cognitive error that best fits the thought \textit{I am responsible for her mistakes} is \textit{Personalization}, as it involves taking personal responsibility for the actions of another person.
}
\end{tcolorbox}

\begin{tcolorbox}[width=0.98\textwidth]
\textbf{Response of Mindora w/o S2 (i.e. the unstructured version):}\\[2pt]
{\small
<think>\\
Okay, so the person is feeling responsible for their sister's mistake. That's interesting because they're taking on this blame for something that wasn't really their fault. It's like they're saying, \textit{I should have known better}, or \textit{It's my fault she didn't register}. \\
Hmm, this sounds like they're taking things personally, right? They're letting this one mistake by their sister affect how they see themselves. It's almost like they're saying, \textit{I'm not a good sibling because I didn't stop her from making this mistake}. \\
But wait, let's think about this a bit more. They're not just blaming themselves; they're also saying they're responsible for the mistake. That's a bit different. It's not just about feeling bad about themselves; it's about thinking they should have done something to prevent the mistake. \\
Oh, this could be \textit{Should statements}. They might be thinking, \textit{I should have made sure she registered}, or \textit{I should have known she'd forget}. It's like they're setting a high bar for themselves and feeling guilty when they don't meet it. \\
But hold on, there's more to it. They're not just setting expectations; they're also feeling responsible for the outcome. That's a bit more about taking ownership of the situation, even though they didn't actually cause the mistake. \\
Hmm, maybe this is \textit{Personalization}. They're making the mistake their own problem, even though it was their sister's action. It's like they're internalizing the blame, thinking it's their fault even though it's not. \\
Alright, so after thinking it through, it seems like \textit{Personalization} fits best. They're taking the mistake and making it about themselves, even though they didn't do anything wrong. It's about how they're letting this one event define their sense of responsibility.\\
</think>\\
<answer>Answer: personalization</answer>
}
\end{tcolorbox}

\caption{Reasoning trajectory comparison between Mindora and Mindora w/o S2.}
\label{fig:s2_ablation}
\end{figure*}

The final Mindora model produces a concise, modular explanation that explicitly isolates the target thought (“I am responsible for her mistakes”), directly maps it to the definition of personalization (misattributing external responsibility to oneself), and terminates with a brief conclusion that tightly links evidence to the predicted label. This organization yields clear step boundaries and minimal compressible content, making the trajectory straightforward to imitate and evaluate.

In contrast, the unstructured trace contains substantial exploratory artifacts (e.g., hedging, self-questioning, and backtracking). The model introduces alternative hypotheses such as should statements, then revises its decision after several paragraphs of reflection. Although the final answer is correct, much of the intermediate text is pedagogically ambiguous: it mixes tentative reasoning with narrative fillers (“Hmm”, “But wait”), and delays the decisive criterion for personalization until late in the chain. For distillation into smaller models, such traces can be suboptimal supervision signals, since the student may learn stylistic verbosity or heuristic option-surfing rather than a stable decision template grounded in the cognitive-error definition.

This example highlights the functional role of S2: it converts exploratory, non-linear reasoning traces into standardized, decision-aligned trajectories with explicit modular structure and a clear evidence-to-label mapping. As a result, S2 improves not only human readability but also the quality of supervision available for training compact models under limited capacity.

\clearpage
\begin{strip}
\centering
\begin{minipage}{0.98\textwidth}
\begin{tcolorbox}[width=\textwidth]
\textbf{DepSign Case Content:}\\[2pt]
{\small
Question: Please choose the most proper depression degree of this poster from: 0,1,2. Where 0 for not depressed, 1 for moderately depressed, and 2 for severely depressed.\\
Context: \\
I don't see any point being here. I'm not going to kill myself, so don't worry about that.
It's just that I don't see any point in me existing. Whenever I look at my peers, they're always happy, have an abundance of friends with them and have this aura of confidence and charisma that attracts anyone. For me, it's the direct opposite; I have a limited amount of friends (or acquaintances more likely), I don't feel happy most of the time and have 0 charisma. Granted, I'm watching videos on how to gain charisma and be more likeable but the progress of it is very gradual. And added with my family and the dynamics of it being fucked up, slogging through college for a degree that I both don't want and not good at (even if I eventually cared about it) and not feeling happy, I shouldn't be here. I honestly feel that I'm taking someone's space in college and they would be better privileged to have it rather than me. It's something that I had on my mind for a while and wanted to put it out there somewhere. Even if it sounds bad grammatically or something like that, I just wanted to get it off my chest.
}
\end{tcolorbox}
\captionof{figure}{Reasoning trajectory comparison between Mindora and Mindora w/o S3 (1).}
\label{fig:s3_ablation1}
\end{minipage}
\end{strip}

\subsection{Consistency Reward (S3).}
We analyze {S3 (consistency reward)} by comparing the full {Mindora} model with {Mindora w/o S3}. Removing S3 leads to a measurable degradation in overall performance: $\mathrm{Avg}_{\mathrm{all}}$ decreases from $0.6933$ to $0.6788$ ($-0.0145$). The effect is most salient on verification- and abstraction-style tasks that require stable evidence integration, where {PSRS} drops from $0.8379$ to $0.6948$ ($-0.1431$) and {MentalMisinfo} decreases from $0.7178$ to $0.6796$ ($-0.0382$).

Trajectory-level evaluation further confirms that S3 improves reasoning reliability. Without S3, {Logical Coherence} decreases ($-0.0583$) and {Informativeness} decreases ($-0.0667$). Usability-related metrics also decline, including {Step Adequacy} ($-0.0375$) and {Terminology Economy} ($-0.0292$). In addition, the {Irrelevance Rate} increases ($+0.0037$), indicating a higher fraction of weakly contributive steps.

Overall, these results suggest that S3 contributes beyond marginal accuracy gains: it promotes more coherent and informative reasoning trajectories while reducing irrelevant content, and is particularly beneficial for tasks where internal inconsistency or ungrounded statements can directly compromise reliability.

\noindent\textbf{Case Study of Consistency Reward (S3)}
\label{append:case_ablation_s3}
This case shown in Figure~\ref{fig:s3_ablation1} involves severity grading for depression (label 1 = moderate). As is shown in Figure~\ref{fig:s3_ablation2}, both Mindora and Mindora w/o S3 reach the same final answer, but their reasoning traces differ in how consistently they connect evidence to the severity decision.

The trajectory of Mindora (with S3) exhibits a clearer evidence-to-decision alignment by explicitly separating risk exclusion (“not going to kill themselves”) from severity evidence (“lack of purpose,” “taking up space”), and grounding the severity level in a combined rationale: moderate depressive features are present, while an acute self-harm signal is absent. The final conclusion explicitly composes these two premises into a severity judgment, making the chain internally well-anchored and decision-oriented.

% The ablated model is largely similar but shows a subtly weaker internal calibration: it introduces an additional criterion (“not accompanied by extreme despair … that would suggest severe depression”) that is not directly supported by the input and functions more like a generic severity template than an evidence-driven inference. In other words, the reasoning partially shifts from text-grounded justification to schema-driven restatement (“common stressors … do not necessarily indicate severe depression”), which increases the risk of over-generalization in harder cases where the boundary between moderate and severe depends on specific clinical markers (e.g., persistence, functional impairment, psychomotor changes, explicit self-harm planning).

% While the difference is small in this example because both outputs remain coherent and reach the correct label, it illustrates the intended role of the consistency reward: discouraging generic, loosely grounded severity templates and encouraging step-wise conclusions that remain tightly coupled to the observed evidence. This aligns with the aggregate findings that S3 improves trajectory reliability and reduces weakly contributive reasoning units, even when final-answer accuracy is unchanged.

The ablated model is similar but less well calibrated: it introduces an unsupported criterion (“not accompanied by extreme despair … that would suggest severe depression”), shifting from text-grounded justification to a generic severity template. Although both outputs remain coherent and reach the correct label, this case illustrates the role of the consistency reward: discouraging loosely grounded templates and encouraging conclusions that stay tightly coupled to observed evidence. This supports the aggregate finding that S3 improves trajectory reliability even when final-answer accuracy is unchanged.

% DepSign_test_119
\begin{figure*}[t]
\centering
\begin{tcolorbox}[width=0.98\textwidth]
\textbf{Response of Mindora:}\\[2pt]
{\small
<think>\\
\#\#\#Evaluating Suicidal Ideation: \\
The poster explicitly states they are not going to kill themselves, indicating a lack of immediate suicidal intent. This suggests they are not in a state of severe depression where self-harm is considered.\\
\#\#\#Analyzing Feelings of Hopelessness: \\
The poster expresses a lack of purpose and feeling like they are taking up space in college. These feelings indicate a level of hopelessness and self-doubt, which are common in moderate depression.\\
\#\#\#Assessing Social and Academic Stress: \\
The poster mentions social isolation and academic struggles, which contribute to feelings of inadequacy. These factors can lead to moderate depression, as they are significant stressors but not necessarily severe.\\
\#\#\#Final Conclusion: \\
The poster's feelings of hopelessness and social inadequacy, combined with their lack of suicidal intent, suggest a moderate level of depression. Therefore, the most appropriate degree of depression for this poster is 1, indicating moderately depressed.\\
</think>
\\
<answer>Answer: 1</answer>
}
\end{tcolorbox}

\begin{tcolorbox}[width=0.98\textwidth]
\textbf{Response of Mindora w/o S3 (i.e. without consistency reward):}\\[2pt]
{\small
<think>\\
\#\#\#Assessing Suicidal Ideation: \\
The poster explicitly states they are not going to kill themselves, indicating a lack of immediate suicidal intent. This suggests they are not in a state of severe depression where suicide is a consideration.\\
\#\#\#Evaluating Emotional State: \\
The poster expresses feelings of hopelessness and inadequacy, which are indicative of depression. However, these feelings are not accompanied by extreme despair or hopelessness that would suggest severe depression.\\
\#\#\#Analyzing Social and Academic Stress: 
The poster mentions social isolation and academic stress, which contribute to their feelings of inadequacy. These are common stressors that can lead to moderate depression but do not necessarily indicate severe depression.\\
\#\#\#Final Conclusion: \\
The poster's expression of hopelessness and inadequacy, combined with their lack of suicidal intent and common stressors, suggests a moderate level of depression. Therefore, the most appropriate depression degree for this poster is 1, indicating moderately depressed.\\
</think>
\\
<answer>Answer: 1</answer>
}
\end{tcolorbox}

\caption{Reasoning trajectory comparison between Mindora and Mindora w/o S3 (2).}
\label{fig:s3_ablation2}
\end{figure*}

\clearpage

\begin{strip}
\centering
\renewcommand{\arraystretch}{1.3}
\captionof{table}{Few-shot in-context learning results on representative \textsc{MiraMind} datasets. CR = CognitiveReframing; PR = PatternReframe.}
\label{tab:icl_analysis}
\resizebox{\linewidth}{!}{
\begin{tabular}{lccccccccccc}
\toprule
Model & Shot & CR & PR & TherapistQA & DepSign & SWMH & T-SID & MHQA & MedQA & MedMCQA & Misinfo \\
\midrule
Qwen3-8B & 0-shot & 0.5941 & 0.6395 & 0.4057 & 0.4169 & 0.7358 & 0.7515 & 0.4009 & 0.5824 & 0.7927 & 0.4773 \\
         & 1-shot & 0.7434 & 0.6111 & 0.4896 & 0.4000 & 0.7541 & 0.8247 & 0.6105 & 0.8020 & 0.8651 & 0.6997 \\
         & 3-shot & 0.7461 & 0.6364 & 0.4667 & 0.4106 & 0.7679 & 0.8305 & 0.6091 & 0.8195 & 0.8808 & 0.7048 \\
\midrule
GPT-o4-mini & 0-shot & 0.7002 & 0.6607 & 0.4468 & 0.3607 & 0.7374 & 0.7515 & 0.5462 & 0.8925 & 0.8102 & 0.7781 \\
            & 1-shot & 0.8906 & 0.6842 & 0.5473 & 0.3673 & 0.7923 & 0.8114 & 0.7127 & 0.9524 & 0.9431 & 0.5167 \\
            & 3-shot & 0.8571 & 0.6696 & 0.5738 & 0.3827 & 0.7805 & 0.8218 & 0.7219 & 0.9386 & 0.9543 & 0.4982 \\
\bottomrule
\end{tabular}
}
\end{strip}

\subsection{Few-shot In-Context Learning Analysis}
\label{append:icl_analysis}

To further examine whether prompt-level adaptation can substitute for post-training, we conduct a small few-shot in-context learning (ICL) analysis following the example-selection paradigm of Medprompt~\citep{nori2023can}. We evaluate GPT-o4-mini and Qwen3-8B under 0-shot, 1-shot, and 3-shot settings on representative \BENCH{} datasets. Table~\ref{tab:icl_analysis} reports the results.

The results show that ICL can improve some task scores, especially on structured classification and QA datasets. For example, Qwen3-8B improves substantially on CognitiveReframing, MHQA, MedQA, MedMCQA, and Misinfo under few-shot prompting, while GPT-o4-mini also benefits on several classification and QA tasks. However, the gains are not consistent across datasets. For instance, Qwen3-8B decreases on PatternReframe and DepSign under 1-shot prompting, and GPT-o4-mini shows a large drop on Misinfo under both 1-shot and 3-shot settings. These mixed results suggest that few-shot prompting is useful as a prompt-level adaptation strategy, but it is insufficient as a general solution for robust mental-health reasoning. In contrast, Mindora obtains more consistent gains through targeted post-training, supporting the necessity of our SFT--RL optimization and reasoning-trajectory design.

\subsection{Trajectory Evaluation Conditioned on Correct Predictions}
\label{append:correct_conditioned_trajectory}

\end{document}